\documentclass[10pt,journal,compsoc]{IEEEtran}
\usepackage{ragged2e}

\ifCLASSOPTIONcompsoc
  \usepackage[nocompress]{cite}
\else
  \usepackage{cite}
\fi
\ifCLASSINFOpdf
\else
\fi

\usepackage{graphicx}
\usepackage{amsmath,amssymb} 
\usepackage{color}
\usepackage[linesnumbered,ruled]{algorithm2e}

\usepackage{caption}
\usepackage{subcaption}
\usepackage{booktabs}
\usepackage{amssymb}
\usepackage{tabularx}
\usepackage{multirow}
\newcolumntype{C}[1]{>{\centering\arraybackslash}p{#1}}

\newif\ifshowcomments
\showcommentstrue

\newcommand{\para}[1]{\vspace{.05in}\noindent\textbf{#1}}
\def\ie{\emph{i.e.}}
\def\eg{\emph{e.g.}}
\def\etal{{\em et al.}}
\def\etc{{\em etc.}}
\newcommand{\new}[1]{{\color{black}#1}}

\begin{document}
%
%
\title{A Rotation-Invariant Framework for \\ Deep Point Cloud Analysis}

%
%
%
%

\author{Xianzhi Li,
	Ruihui Li,
	Guangyong Chen,
	Chi-Wing~Fu,
	Daniel Cohen-Or
	and~Pheng-Ann~Heng
	\IEEEcompsocitemizethanks{\IEEEcompsocthanksitem X. Li, R. Li, C.-W. Fu, and P.-A. Heng are with the Chinese University of Hong Kong. \protect
		E-mail: \{xzli, lirh, cwfu, pheng\}@cse.cuhk.edu.hk
	
		G. Chen and P.-A. Heng is also with Shenzhen Key Laboratory of Virtual Reality and Human Interaction Technology, Shenzhen Institutes of Advanced Technology, Chinese Academy of Sciences, China.
		E-mail: gy.chen@siat.ac.cn

	D. Cohen-Or is with Tel Aviv University.
		E-mail: dcor@mail.tau.ac.il
		
	R. Li is the corresponding author.
	}
}

%
%

\markboth{IEEE Transactions on Visualization and Computer Graphics}
{(under review)}
\IEEEtitleabstractindextext{%

\begin{justify}
\begin{abstract}
Recently, many deep neural networks were designed to process 3D point clouds, but a common drawback is that rotation invariance is not ensured, leading to poor generalization to arbitrary orientations.
In this paper, we introduce a new low-level purely rotation-invariant representation to replace common 3D Cartesian coordinates as the network inputs.
Also, we present a network architecture to embed these representations into features, encoding local relations between points and their neighbors, and the global shape structure.
To alleviate inevitable global information loss caused by the rotation-invariant representations, we further introduce a region relation convolution to encode local and non-local information.
We evaluate our method on multiple point cloud analysis tasks, including (i) shape classification, (ii) part segmentation, and (iii) shape retrieval.
Extensive experimental results show that our method achieves consistent, and also the best performance, on inputs at arbitrary orientations, compared with all the state-of-the-art methods.
\end{abstract}
\end{justify}

}

\maketitle

\IEEEdisplaynontitleabstractindextext

%
\IEEEpeerreviewmaketitle

\section{Introduction}
The development of neural networks for point cloud analysis has drawn a lot of interests in recent years,
and applied to various 3D applications,~\eg, shape classification, object detection, semantic scene segmentation, etc.
However, the features learned by most existing networks are \emph{not\/} rotation invariant, meaning that they regard a point cloud and an arbitrary rotation of it as two different shapes. Thus, different features could be extracted from the same shape, that is merely embedded in different
poses in 3D.

\begin{figure}[t]
	\centering
	\includegraphics[width=0.93\linewidth]{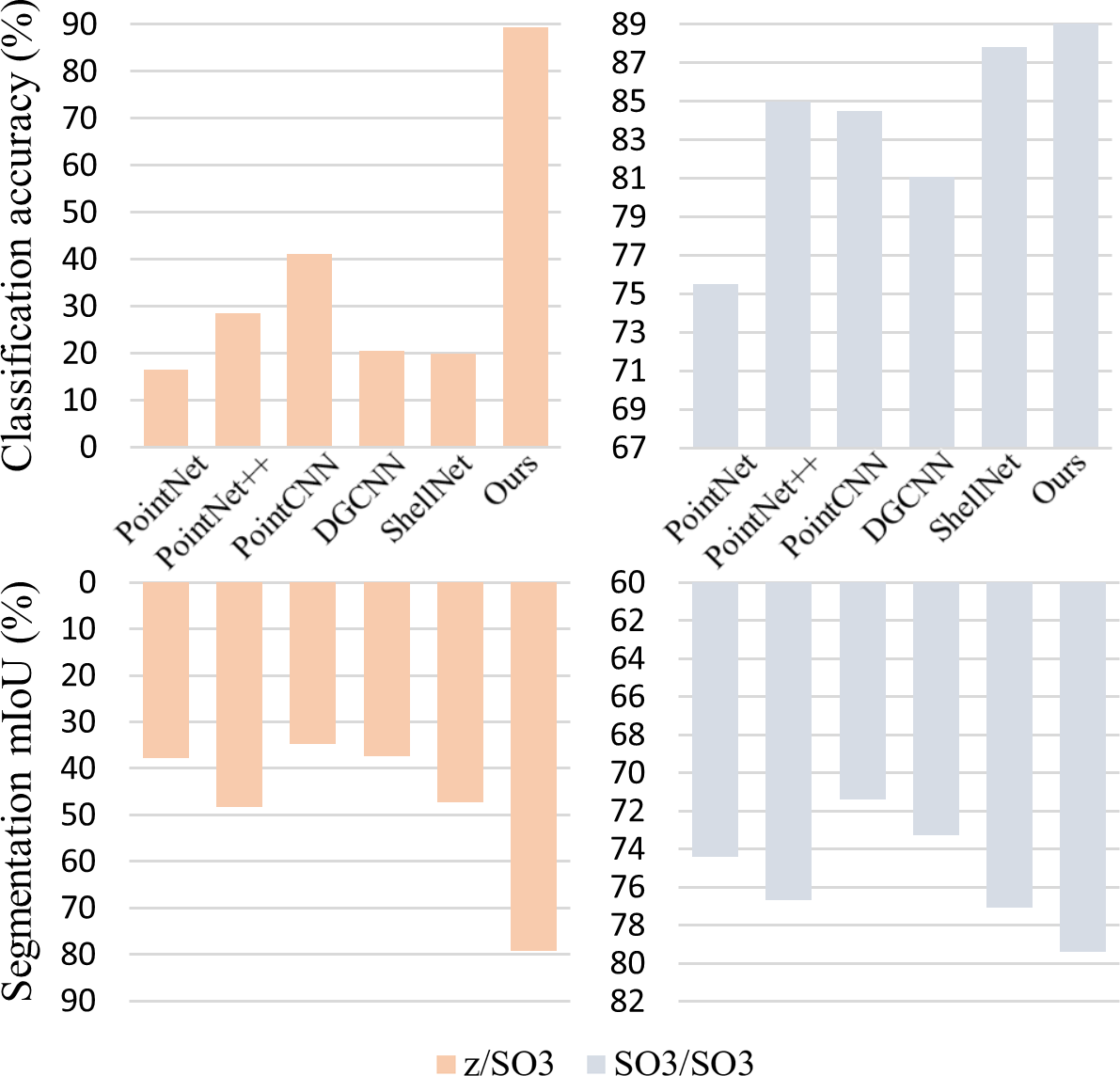}
	\caption{Performance comparisons on shape classification (top) and part segmentation (bottom) for two scenarios: z/SO3 and SO3/SO3.
	When trained in azimuthal rotations and tested in arbitrary rotations (z/SO3), the rotation-variant methods (e.g., PointNet, etc.) have poor performance, while our method clearly outperforms them; see the orange bars on the left.
	On the other hand, even trained and tested both in arbitrary rotations (SO3/SO3), our method still achieves the best performance; see the gray bars on the right.	
	Section~\ref{sec:experiments} contains more experimental results and quantitative measures on the shape retrieval task.}
	\label{fig:teaser}
\vspace{-2mm}
\end{figure}

To alleviate this fundamental problem, a common approach is to apply rotation augmentation to the training data. However, aggressive rotation augmentation, like arbitrary 3D rotations,
often harms the recognition performance, since most existing networks do not have strong capacity to learn effective features from such unstable inputs. Thus, often, only azimuthal rotations (around the gravity axis) are considered during network training. However, such limited augmentation does not generalize well, which could lead to a very poor performance given inputs on arbitrary rotations. See Figure~\ref{fig:teaser},~\eg, the classification accuracy of DGCNN~\cite{wang2018dynamic} and the recent ShellNet~\cite{zhang2019shellnet} can only achieve around 20\% (top left orange bar), when trained in azimuthal rotations and tested in arbitrary rotations (denoted as z/SO3).

Recently, some works attempt to design neural networks with rotation invariance~\cite{poulenard2019effective,rao2019spherical,deng2018ppf,zhang2019rotation,chen2019clusternet}.
One approach employs spherical-related convolutions, and the other employs
local rotation-invariant features,~\eg, distances and angles,
to replace Cartesian coordinates as the network inputs.
However, as we shall show, both approaches have limited success.
Typically, the former approach has limited capability to embed features and is still sensitive to the rotations, while the latter one encodes mainly local information, which may not be unique, and often, the performance is much lower than using global Cartesian coordinates;
see various comparison results in Section~\ref{sec:experiments}, which demonstrate the superiority of our method over the state-of-the-art rotation-invariant methods.

In this work, we revisit the problem of rotation invariance in deep 3D point cloud analysis, and enumerate the {\em considerations for achieving rotation invariance\/} in the aspects of network inputs and network processing.
Accordingly,
\begin{itemize}
	\item
	We first design an effective low-level representation to replace 3D Cartesian coordinates as the network inputs.
	Our representation is {\em purely rotation-invariant\/} and encodes {\em both local and global\/} information, as well as being {\em robust to noise and outliers\/}.
	\vspace*{2mm}
	\item
	We present a {\em deep hierarchical network\/} to embed these low-level representations into high-level features and extract local relations between points and their neighbors, together with global shape information.
		\vspace*{2mm}
	\item
	Further, to alleviate global information loss caused by the rotation-invariant representations,
	we enrich the network features with more global information by introducing a novel {\em region relation convolution\/} in the network to extract both local and non-local information across the hierarchy.
\end{itemize}

Lastly, we evaluate the effectiveness of our method on various point cloud analysis tasks, including shape classification, part segmentation, and shape retrieval. Experimental results confirm that, our method achieves not only consistent results on inputs at any orientation, but also the best performance on all tasks compared with state-of-the-arts.

\section{Related Works}
\label{sec:rw}


\para{Deep learning on 3D point sets.} \
The design of robust and effective neural networks to embed point features has been an emerging topic
in recent years.
The pioneering networks PointNet~\cite{qi2017pointnet} and PointNet++~\cite{qi2017pointnet++} show the potentials of deep networks to directly process 3D point sets.
%
To better capture local neighborhoods, several works~\cite{wang2018dynamic,shen2018mining,liu2019relation,zhao2019pointweb}
suggest extracting point features by considering local graphs.
To address the irregular and orderless properties of point sets, some works define convolutions on
non-Euclidean domains,~\eg,
the self-organizing map~\cite{li2018so}, $\mathcal{X}$-transformation~\cite{li2018pointcnn}, permutohedral lattice~\cite{su2018splatnet}, and parameterized embedding~\cite{xu2018spidercnn}.
Some others propose new convolution operators on points,~\eg, Monte Carlo convolution~\cite{hermosilla2018monte}, PointConv~\cite{wu2019pointconv}, ShellConv~\cite{zhang2019shellnet}, and KPConv~\cite{thomas2019kpconv}.

Apart from learning point features in a supervised manner, some unsupervised and self-supervised networks~\cite{yang2018foldingnet,zhao20193d,chen2019deep,han2019multi,sauder2019self,hassani2019unsupervise,li2020unsupervised,fernandez2020unsupervised} were designed recently to avoid tedious manual labeling.
Besides high-level object recognition and detection, some networks were designed for point set registration~\cite{aoki2019pointnetlk,deep2019registration,deepvcp2019registration}, upsampling~\cite{yu2018pu,yifan2019patch,li2019pu,qian2020pugeo}, completion~\cite{xie2020grnet}, and denoising~\cite{hermosilla2019total,zhou2019defense}, etc.

Although these networks are translation and permutation invariant, they are {\em not\/} rotation invariant.
They embed different features, and likely produce different outputs for the same input
given in different orientations.

\para{Rotation-equivariant networks for 3D shapes.} \
To improve the robustness of the network for rotation, some works attempted to design rotation-equivariant networks, where the learned features rotate correspondingly with the input.
For example, considering that vanilla CNNs only have translation invariance, spherical CNNs~\cite{esteves2018learning,cohen2018spherical}, and 3D steerable CNNs~\cite{weiler20183d} were introduced to learn rotation-equivariant features.
Recently, motivated by the fact that quaternion naturally satisfies the rotation-equivariant property,~\cite{zhao2020quaternion} and~\cite{zhang20203d} proposed quaternion-based neural networks.
However, since the features learned by these methods rotate correspondingly with the input, the analysis results may change subject to rotations on the input, so they may not guarantee stable classification and segmentation results.
Further, both spherical CNNs and quaternion-based networks have limited learning capability, thus leading to a notable performance drop compared with the non-rotation-equivariant methods.
We shall show this in Section~\ref{sec:experiments}.


\para{Rotation-invariant networks for 3D shapes.} \
To ensure consistent shape analysis results given inputs at arbitrary orientations, the fundamental approach is to design rotation-invariant methods.
Recently, some works explored rotation-invariant networks for point clouds.
Poulenard~\etal~\cite{poulenard2019effective} represented points using volume functions, then used spherical harmonics kernels for convolution.
The feature embedding capability of such convolution is, however, limited.
Rao~\etal~\cite{rao2019spherical} adaptively projected points on a discretized sphere and designed a hierarchical feature learning architecture to capture patterns on the sphere.
However, the discretized sphere still carries a global orientation and cannot guarantee perfect symmetry, so the learned features are not purely rotation invariant.
Hence, a notable performance drop still exists for inputs at arbitrary orientations.

On the other hand, some other methods suggested using low-level rotation-invariant geometric features to replace 3D Cartesian coordinates as the network inputs.
Deng~\etal~\cite{deng2018ppf} suggested relative angles between point-wise normal vectors and paired distances.
Chen~\etal~\cite{chen2019clusternet} suggested relative angles between two-point vectors, and vector norm,~\etc\/
Zhang~\etal~\cite{zhang2019rotation} constructed a point's neighborhood with local triangles, each formed by a reference point, a neighbor point, and the local neighborhood centroid.
They then take the triangle side lengths and angles as the rotation-invariant features.
Though these representations are rotation invariant, they encode mainly local information, which may not be unique and sufficient; see Section~\ref{subsec:considerations} and our supplementary material for a detailed analysis.
In this work, we present a new rotation-invariant representation, capturing both local neighborhood and global shape structures, while being robust to noise and outliers.
Also, we formulate a deep network, and introduce a novel region relation convolution to hierarchically process the point regions.

\section{Method}
\label{sec:method}


\subsection{General Model for Point Feature Extraction}
\label{subsec:bg}
To start, we review a general model (w/o rotation invariance) for point feature extraction.
Denote $\mathcal{S}=\{p_1, p_2, \dots,$ $p_N\}$ as a point cloud of $N$ points, where $p_i$ is the 3D Cartesian coordinate of the $i$-th point in $\mathcal{S}$.
To extract features for a point, say $p_i$, a general model would include both {\em local\/} and {\em global\/} information, so can be written as
\begin{equation}
\label{equ:convolution}
\mathcal{A}_{j=1}^K(h_{\theta}(G_i,L_{ij})),
\end{equation}
where
$G_i$ denotes the global shape information at $p_i$;
$L_{ij}$ denotes the local shape information at $p_i$ with its $j$-th neighbor point $p_{ij}$ ($j = 1..K$);
$h_{\theta}$ is a nonlinear function with learnable parameters $\theta$; and
$\mathcal{A}$ is a symmetric aggregation operation for point permutation invariance,~\eg, max or summation, over the $K$ neighbor points of $p_i$.

For general points processing networks without considering rotation invariance, $G_i$ is simply represented by $p_i$, since 3D coordinates are global.
For $L_{ij}$, different networks have different choices,~\eg,
PointNet++~\cite{qi2017pointnet++} uses $p_{ij}$ as $L_{ij}$, while DGCNN~\cite{wang2018dynamic} uses relative position,~\ie, $p_{ij}-p_i$, as $L_{ij}$.
Clearly, both $G_i$ and $L_{ij}$ are based on 3D coordinates, so they are {\em not\/} rotation invariant.


\subsection{Considerations for Rotation Invariance}
\label{subsec:considerations}

First of all, a framework is rotation invariant, if both the network inputs and operations are rotation invariant.
Hence, before we introduce our network inputs (Section~\ref{subsec:rir}) and network architecture (Section~\ref{subsec:network}), we first discuss the relevant design considerations that we have taken:

\vspace*{2mm}
\para{Considerations for designing the network inputs.} \
\begin{itemize}
%
\item[(i)]
The fundamental rule we should follow to design the network inputs is that, no matter how the input point cloud $\mathcal{S}$ rotates, the extracted rotation-invariant representations (network inputs) must remain unchanged.
Mathematically, denoting $\Phi$ as the function to extract network inputs from point cloud $\mathcal{S}$, a {\em purely rotation-invariant\/} $\Phi$ should satisfy
\begin{equation}
\label{equ:purely}
\Phi(\mathcal{S}) = \Phi(R(\mathcal{S})),
\end{equation}
where $R \in$ SO(3)\footnote{SO(3) is the space of all 3D rotations in $\mathbb{R}^3$.} is an arbitrary rotation.

\vspace*{1mm}
\item[(ii)]
Second, to design $\Phi$ that satisfies Eq.~\eqref{equ:purely}, we may simply use $L_2$ distances or relative angles between nearby point pairs as the rotation-invariant network inputs.
However, using just local relative geometric quantities may cause a large amount of information loss and lead to a considerable performance drop compared with the non-rotation-invariant methods.
Hence, we shall try to design rotation-invariant network inputs that can retain as much useful (shape) information as possible.

\vspace*{1mm}
\item[(iii)]
\new{Third, the designed rotation-invariant network inputs should be \emph{ambiguity-free}, meaning that different local regions have their own unique rotation-invariant representations.
Otherwise, it may cause ambiguity in the network learning.}

\vspace*{1mm}
\item[(iv)]
Last, noise is often unavoidable when scanning 3D point clouds.
Hence, $\Phi$ should be \emph{noise tolerant}, meaning that the extracted rotation-invariant representations should not be too sensitive to noise in $\mathcal{S}$.

\end{itemize}

Current attempts~\cite{deng2018ppf,zhang2019rotation,chen2019clusternet} considered only item (i) above by proposing point-wise purely rotation-invariant representations to replace 3D point coordinates as the network inputs.
However, they focus mainly on the encoding of local relations $L_{ij}$ between nearby point pairs using,~\eg, $L_2$ distances and relative angles, and ignore $G_i$, thus leading to information loss.
Also, most existing methods suffer from the ambiguity of distinguishing between local shapes.
Please refer to our supplementary material for detailed discussion.
Particularly, none of them considers the noise-tolerance issue.
In contrast, we design our rotation-invariant representations for both $L_{ij}$ and $G_i$, and also make it noise-tolerant.
Meanwhile, our designed local point-wise rotation-invariant network input is ambiguity-free.

\para{Considerations for designing the network architecture.} \
\begin{itemize}
%
\item[(i)]
First and foremost, the designed network architecture cannot involve any rotation-variant operations. For instance, the network should not assume specific order (which may not be rotation invariant)
when processing/aggregating point and regional features.
Further, many existing networks,~\eg,~\cite{liu2019relation,duan2019structural,chen2019gapnet}, regress attention weights using point coordinates and provide guidance to enhance the embedded features; we cannot follow this practice, since it uses rotation-variant information.

\vspace*{1mm}
\item[(ii)]
A rotation-invariant network should not take point coordinates but only relative geometric information, such as distances and angles, as its inputs.
However, without absolute information defined in a global coordinate frame, the network would lack global information, thus deteriorating the performance of high-level tasks,~\eg, shape classification and retrieval.
Hence,
we should extract more global features, even from the relative geometric inputs, by considering more global relations among points.
Existing rotation-invariant methods did not explore the global point relations, as in our work.
\end{itemize}

\begin{figure*}[t]
	\centering
	\includegraphics[width=0.97\linewidth]{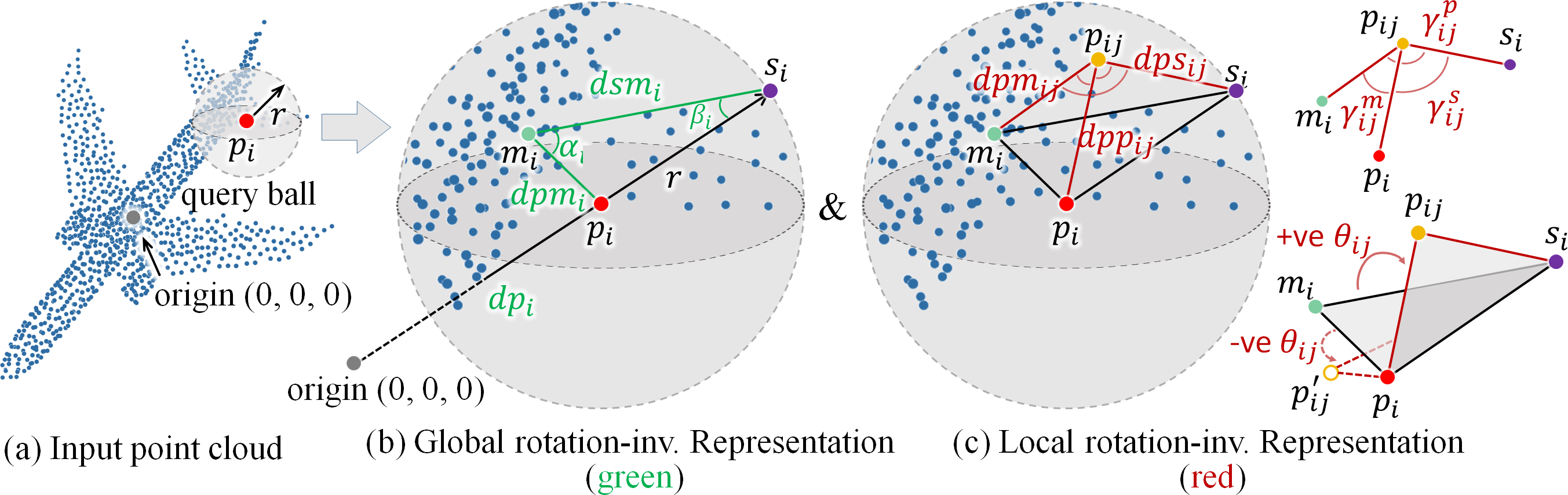}
	\caption{Our global and local rotation-invariant representations extracted at point $p_i$.}
	\label{fig:representation}
\end{figure*}

\vspace*{-1.0mm}
\subsection{Our Rotation-Invariant Representations}
\label{subsec:rir}

Before extracting our rotation-invariant representations for $G_i$ and $L_{ij}$, we first normalize input point cloud $\mathcal{S}$
by fitting it in the origin-centered unit sphere.
Then, for each point $p_i$ in $\mathcal{S}$ (\eg, the red point in Figure~\ref{fig:representation}(a)), we follow PointNet++~\cite{qi2017pointnet++} to use a query ball of radius $r$ to locate $K$ neighbor points $\{p_{ij}\}_{j=1}^K$ (including $p_i$ itself) as its local neighborhood (blue points in Figure~\ref{fig:representation}(b)).

Now, we are ready to extract $G_i$ and $\{ L_{ij} \}_{j=1}^K$ of $p_i$.
On the one hand, $G_i$ should capture $p_i$'s location relative to the whole object and to its local neighborhood, serving {\em like\/} the global but rotation-variant coordinates of $p_i$ in the object.
On the other hand, $\{ L_{ij} \}_{j=1}^K$ of $p_i$ should capture the local shape around $p_i$, so we model a local representation for each $p_{ij}$, serving like the local rotation-variant coordinates of $p_{ij}$ in $p_i$'s local neighborhood.
%


\para{Global representation $G_i$} includes the following five pieces of rotation-invariant information about $p_i$ (see Figure~\ref{fig:representation}(b) for an illustation of these five components).

(i) $dp_i$$=$$\Arrowvert p_i\Arrowvert^2$, simple but global and rotation-invariant information about $p_i$.

(ii) $dpm_i$, the distance from $p_i$ to $p_i$'s local neighborhood center (denoted as $m_i$).
Here, a common choice of $m_i$ is $\{p_{ij}\}$'s centroid (arithmetic mean), but such $m_i$ is sensitive to outliers and noise, so $dpm_i$
may not be stable.
We propose to use the geometric median,~\ie, the point with minimal distance sum to all $\{p_{ij}\}$.
Such a choice is more stable,
but computationally expensive~\cite{cohen2016geometric}.
So, we resort to a fast but approximate procedure based on the idea of \emph{divide and conquer}: we first randomly and independently pick $P_s$ subsets of $K_s$ points in $\{p_{ij}\}$, find the centroid of each subset, and cluster the centroids.
Then, we take the mean of the centroids in the largest cluster as $m_i$.
Please refer to Section~\ref{subsec:implementation} for hyper-parameters $P_s$ and $K_s$, Section~\ref{subsec:noise} for a noise tolerance experiment, and our supplementary material for an evaluation of the approximated $m_i$.

(iii)-(v) we locate $s_i$ (purple point in Figure~\ref{fig:representation}(b)), the intersection between the query ball and line extended from origin to $p_i$, and form triangle $p_i$-$m_i$-$s_i$.
Then, we consider $dsm_i$, the distance from $s_i$ to $m_i$, and the cosine of the two angles subtended at $m_i$ and $s_i$ (denoted as $\alpha_i$ and $\beta_i$) as the last three components of $G_i$; see Figure~\ref{fig:representation}(b).
In our implementation, the query ball radius increases with the network layer (see Section~\ref{subsec:network}), so the underlying structure described by triangle $p_i$-$m_i$-$s_i$ will enlarge gradually, thus being more and more global.

Note that, component (i) roughly represents $p_i$'s location related to the whole object, while components (ii)-(v) capture $p_i$'s reference to its local neighborhood. Compared with the global but rotation-variant coordinates of $p_i$, we design the five components to capture the global rotation-invariant representation of $p_i$ with minimal information loss.
To sum up, our global rotation-invariant representation is
\begin{equation}
\label{equ:center}
G_i = [dp_i, dpm_i, dsm_i, \text{cos}(\alpha_i), \text{cos}(\beta_i)] \ .
\end{equation}
Also, note that all distances range $[0,1]$ (since the input point cloud has been normalized), whereas angles $\alpha_i$ and $\beta_i$ range $[0,\pi]$.
To avoid unstable gradient flow, which may result in a slow or oscillating learning process, we take cosine of these two angles in $G_i$.

\para{Local representation} $L_{ij}$ should help to \emph{uniquely} locate $p_{ij}$ relative to $p_i$ in $p_i$'s local neighborhood.
To achieve rotation invariance, similar to $G_i$, the construction of $L_{ij}$ can only be based on the $L_2$ distances and relative angles.
Naturally, given $p_{ij}$ and triangle $p_i$-$m_i$-$s_i$, we first construct a tetrahedron by joining $p_{ij}$ to triangle $p_i$-$m_i$-$s_i$, and consider the three distances $dpm_{ij}$, $dpp_{ij}$, and $dps_{ij}$ from $p_{ij}$ to points $m_i$, $p_i$, and $s_i$, respectively, and the three angles $\gamma_{ij}^p$, $\gamma_{ij}^m$, and $\gamma_{ij}^s$ subtended at $p_{ij}$ on the three tetrahedron faces; see Figure~\ref{fig:representation}(c).
Using these information alone may be ambiguous, since a mirror point $p'_{ij}$ of $p_{ij}$ on the opposite side of triangle $p_i$-$m_i$-$s_i$ can have the same set of distances and angles; see Figure~\ref{fig:representation}(c).
So, we further consider $\theta_{ij} \in [-\pi, \pi)$, the angle for the rotating plane of triangle $m_i$-$p_i$-$s_i$ to the plane of triangle $p_{ij}$-$s_i$-$p_i$ about line $p_i$-$s_i$.

Again, to avoid unstable gradient flow, we take cosine of $\gamma_{ij}^p$, $\gamma_{ij}^m$, and $\gamma_{ij}^s$.
As for $\theta_{ij}$, since it ranges $[-\pi, \pi)$, we
use a nonlinear function $f(\theta_{ij})$$=$$\text{sin}(\frac{1}{2}\theta_{ij})$, which is monotonic for $\theta_{ij}$$\in$$[-\pi, \pi)$ and also ranges $[-1, 1]$.
To sum up, our ambiguity-free local rotation-invariant representation for point $p_{ij}$ relative to $p_i$ is
\begin{equation}
\begin{aligned}
\label{equ:neighbor_all}
L_{ij} = [&dpm_{ij}, dpp_{ij}, dps_{ij}, \\
&\text{cos}(\gamma_{ij}^p), \text{cos}(\gamma_{ij}^m), \text{cos}(\gamma_{ij}^s), f(\theta_{ij})].
\end{aligned}
\end{equation}
Please refer to supplementary file for the proof on the ambiguity-free property.

\begin{figure}[t]
	\centering
	\includegraphics[width=0.99\linewidth]{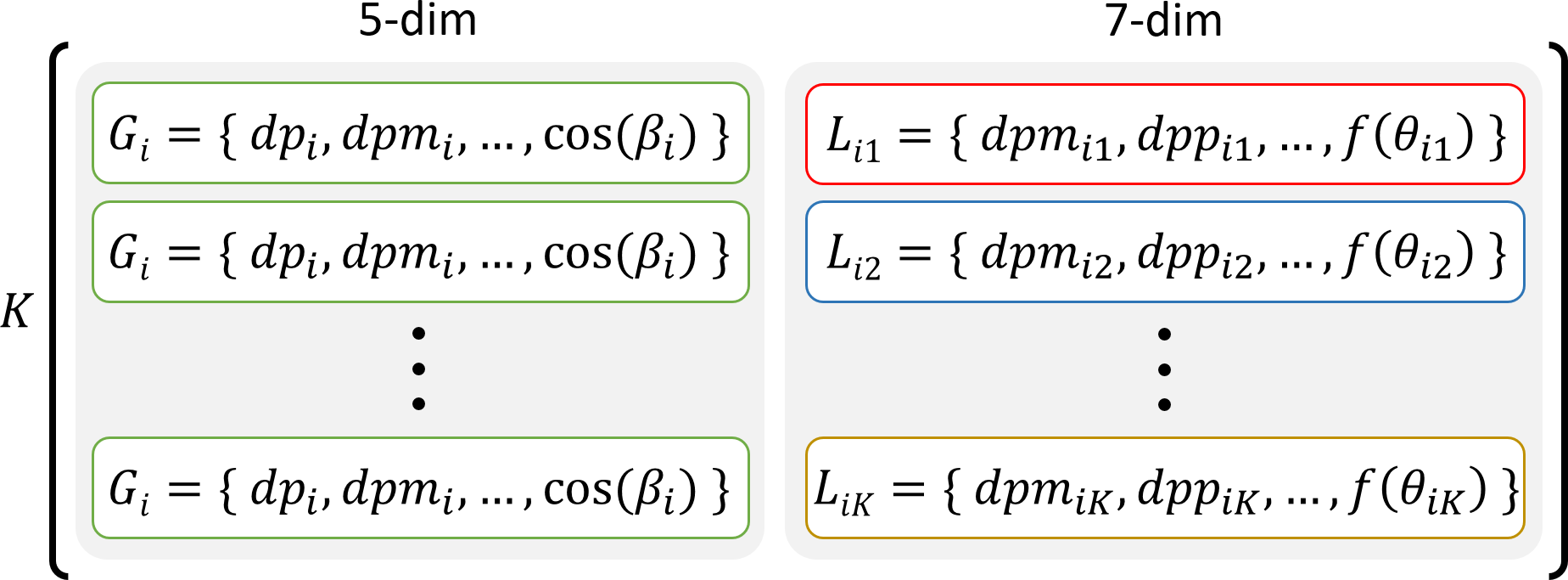}
	\caption{Rotation-invariant network inputs for point $p_i$.}
	\label{fig:matrix}
\end{figure}

\begin{figure*}[t]
	\centering
	\includegraphics[width=0.935\linewidth]{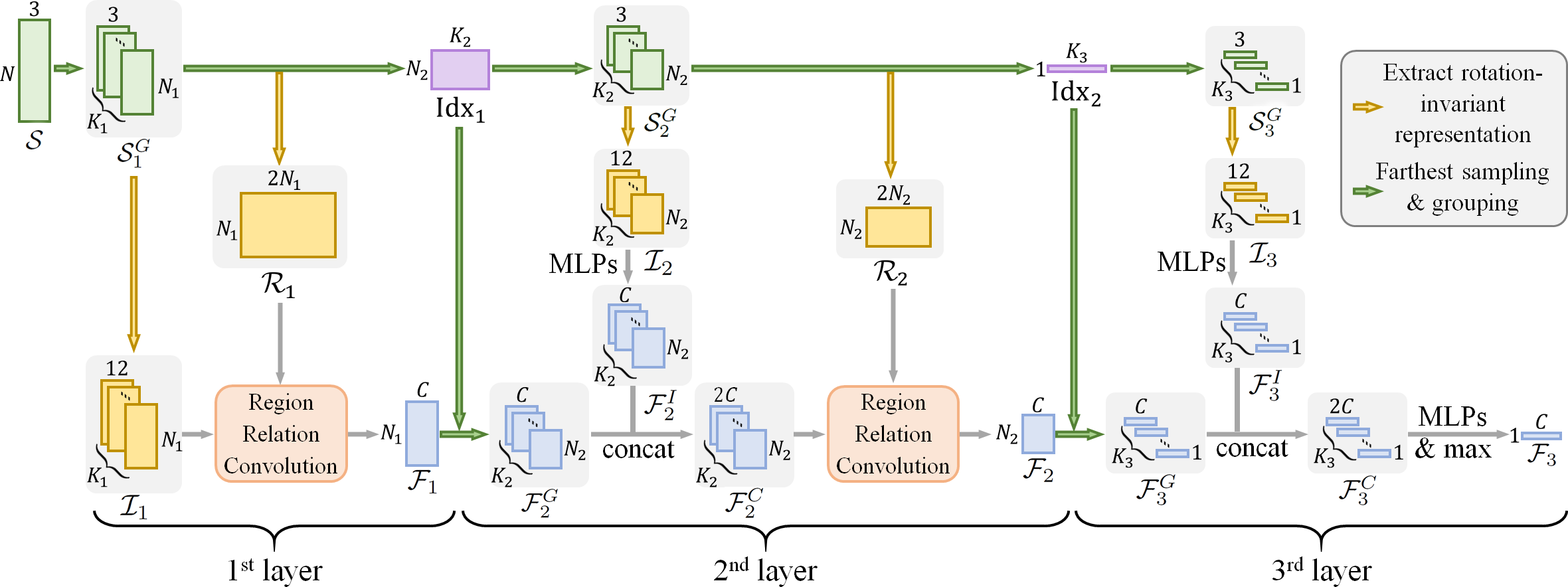}
	\caption{Illustrating the architecture of our deep hierarchical network.
		Given an input point cloud $\mathcal{S}$ ($N$ points), we use farthest sampling \& grouping (green arrows) to introduce enlarging receptive fields in the three-layer hierarchy: $N$$>$$N_1$$>$$N_2$ and $K_1$$<$$K_2$$<$$K_3$.
		From the 3D point coordinates (green boxes), we extract rotation-invariant representations (yellow arrows \& boxes) and pass them to MLPs and region relation convolutions (see Figure~\ref{fig:conv}) to embed features (blue boxes).}
	\label{fig:network}
	\vspace*{-1mm}
\end{figure*}

Overall, for each point $p_i$$\in$$\mathcal{S}$, we employ Eq.~\eqref{equ:center} to obtain its $G_i$ and Eq.~\eqref{equ:neighbor_all} to obtain its $L_{ij}$ (for each of its $K$ neighbor points $\{p_{ij}\}_{j=1}^K$).
Then,
we pack $K$ copies of $G_i$ with $\{L_{ij}\}_{j=1}^K$ to form a $K$$\times$$12$ matrix (see Figure~\ref{fig:matrix}) to store the global and local rotation-invariant representations for $p_i$.
We have to clarify that, although some of the angles in Eqs.~\eqref{equ:center} and \eqref{equ:neighbor_all} seem to be redundant because they can be derived from the distance values, the derivations are not entirely straight-forward.
Without providing these angles, the network might have to learn to compute them all by itself, leading to extra workload.
Also, distances and angles reflect different aspects of geometric information, thus including them both in the network input gives the network higher flexibility to learn the geometric information.
Our experimental results also confirm that if we remove these angles, the shape classification accuracy on ModelNet40 dataset will drop from 89.4\% to 88.1\%.

\vspace*{1mm}
\para{Global relation between points.} \
To supplement $G_i$ and $L_{ij}$ with more global and rotation-invariant information, we further construct $\mathcal{R}$, an $N \times 2N$ matrix to encode the global relations between all point pairs in a point cloud (say, of $N$ points), where matrix elements $\mathcal{R}_{i,2j-1}$ and $\mathcal{R}_{i,2j}$ encode the distance between $p_i$ and $p_j$, and the angle between the two vectors from the origin to $p_i$ and $p_j$, respectively.
These information are later fed into the region relation convolution in the network to regress the rotation-invariant point-wise relation weights; see Section~\ref{subsec:network}.


\subsection{Network Architecture}
\label{subsec:network}
Guided by the considerations presented in Section~\ref{subsec:considerations}, we design a deep hierarchical network of three layers to embed a rotation-invariant codeword of the input point cloud.
Figure~\ref{fig:network} illustrates the network architecture, where the green boxes denote 3D point coordinates (\eg, $\mathcal{S}$) sampled from the input point cloud;
yellow boxes denote extracted rotation-invariant representations (see Section~\ref{subsec:rir});
purple boxes denote point indices
from
farthest sampling; and
blue boxes denote embedded features in the network.

Specifically, given a point set $\mathcal{S}$ of $N$ points, like PointNet++~\cite{qi2017pointnet++}, we first adopt a sample-and-group operator.
That is, we use farthest sampling to select a subset of $N_1$ points, then for each sampled point, we use a query ball to find its $K_1$ neighbor points and group an $N_1$$\times$$K_1$$\times$$3$ volume of 3D point coordinates; see $\mathcal{S}_1^G$ in Figure~\ref{fig:network}.
We then follow the steps in Section~\ref{subsec:rir} to map it into our rotation-invariant representations $\mathcal{I}_1$$\in$$\mathbb{R}^{N_1 \times K_1 \times 12}$ (yellow box) and compute an $N_1$$\times$$2N_1$ global relation matrix $\mathcal{R}_1$ (yellow box) from the sampled $N_1$ points.
Further, we feed $\mathcal{I}_1$ and $\mathcal{R}_1$ into the region relation convolution (to be presented later) to obtain the feature map $\mathcal{F}_1$$\in$$\mathbb{R}^{N_1 \times C}$ (blue box) of the first layer.

The second layer continues to sample-and-group $\mathcal{S}_1^G$ into a smaller point subset $\mathcal{S}_2^G \in \mathbb{R}^{N_2 \times K_2 \times 3}$ and uses the same set of indices (Idx$_1$) to group $\mathcal{F}_1$ into $\mathcal{F}_2^G \in \mathbb{R}^{N_2 \times K_2 \times C}$.
Note that, we set $N_2$$<$$N_1$ and $K_2$$>$$K_1$ to allow a progressively enlarging receptive field in the hierarchy.
Instead of directly feeding $\mathcal{F}_2^G$ into region relation convolution for feature embedding, we avoid information loss by concatenating $\mathcal{F}_2^G$ and $\mathcal{F}_2^I \in \mathbb{R}^{N_2 \times K_2 \times C}$, which are high-level features extracted from low-level representations $\mathcal{I}_2$ via a series of multi-layer perceptron (MLPs); see Figure~\ref{fig:network}.
We then feed the concatenated features $\mathcal{F}_2^C \in \mathbb{R}^{N_2 \times K_2 \times 2C}$, together with another global relation matrix $\mathcal{R}_2 \in \mathbb{R}^{N_2 \times 2N_2}$ from $\mathcal{S}_2^G$, to another region relation convolution to generate $\mathcal{F}_2$ as the output from the second layer.

\begin{figure}[t]
	\centering
	\includegraphics[width=0.99\linewidth]{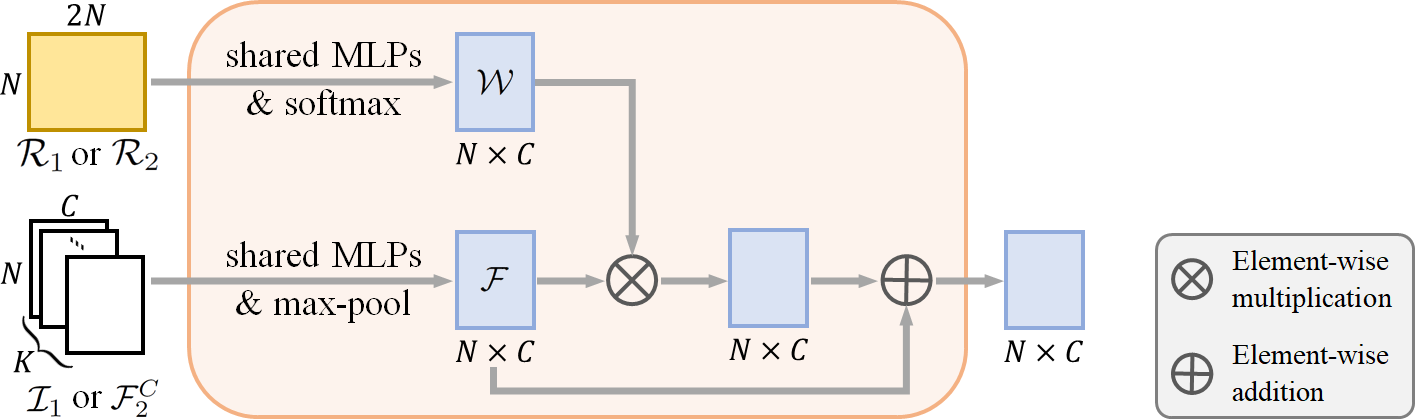}
	\caption{Region relation convolution takes global relations ($\mathcal{R}_1$ or $\mathcal{R}_2$) between points (see Section~\ref{subsec:rir}) to introduce more global information into the deep neural network.
		Note, $N$ in the above illustration is $N_1$ or $N_2$ and $K$ is $K_1$ or $K_2$, depending on the 1$^{\text{st}}$ or 2$^{\text{nd}}$ network layer (see Figure~\ref{fig:network}).}
	\label{fig:conv}
\end{figure}

Further, the third layer samples-and-groups $\mathcal{S}_2^G$ into $\mathcal{S}_3^G \in \mathbb{R}^{1 \times K_3 \times 3}$, and uses the concatenated features $\mathcal{F}_3^C \in \mathbb{R}^{1 \times K_3 \times 2C}$ for convolution.
Now, we only have one single point together with its $K_3$ neighbors ($K_3$$>$$K_2$), so we directly use MLPs followed by max-pooling along $K_3$ on $\mathcal{F}_3^C$ to produce the global feature vector $\mathcal{F}_3 \in \mathbb{R}^{1 \times C}$, which is a rotation-invariant codeword of the input point set.

Next, we can use $\mathcal{F}_3$ in various point cloud analysis tasks.
For examples, for shape classification, we can follow the common routine of using fully-connected layers to regress the class scores.
For part segmentation, we can adopt the point feature propagation and interpolation~\cite{qi2017pointnet++} to recover the per-point features, then
use MLPs to regress per-point scores; please see~\cite{qi2017pointnet++} for details.
For shape retrieval, we can directly compare the cosine similarity between the codewords of the query and target point clouds.
For all the tasks, we use the standard cross entropy loss to guide the network learning.

\para{Region relation convolution.} \
To alleviate the inevitable global information loss in the rotation-invariant representations, we further formulate the region relation convolution (see Figure~\ref{fig:conv} for its illustration) to regress global region relation weight $\mathcal{W}$ from the global relation matrix $\mathcal{R}$ (which is $\mathcal{R}_1$ or $\mathcal{R}_2$) and to refine feature $\mathcal{F}$ extracted from $\mathcal{I}_1$ or $\mathcal{F}^C_2$.
Here, for each reference point and its $K$ local neighbors,
previous networks~\cite{qi2017pointnet++,wang2018dynamic} commonly apply shared MLPs to the $K$ point features and max-pooling along $K$ to obtain a $1$$\times$$C$ feature vector for encoding the local structure around the reference point.
The same operation is applied to all $N$ points to obtain an $N$$\times$$C$ feature map $\mathcal{F}$.
Such operation, however, considers only $p_i$'s own local region when extracting point features for $p_i$, without looking at its relations with other points more globally.

To introduce more global information into the embedded features,
compared with conventional convolutions~\cite{qi2017pointnet++,wang2018dynamic}, after the shared MLPs and max-pooling, we refine features $\mathcal{F}$ by regressing rotation-invariant region relation weight $\mathcal{W}$ from the global relation matrix $\mathcal{R}$; see the top branch in Figure~\ref{fig:conv}.
The weights in each row, say $\mathcal{W}_i$, are regressed based on distances and angles of $p_i$ relative to all the other points (see the last paragraph in Section~\ref{subsec:rir} for details), so $\mathcal{W}_i$ reveals certain global relations between $p_i$ and other points.
We then bring such global information into $\mathcal{F}$ by $\mathcal{F} \oplus (\mathcal{W}$$\otimes$$\mathcal{F}$), where
$\oplus$ and $\otimes$ mean element-wise addition and multiplication, respectively.
Hence, the features of each point encode not only the local structure around the associated point, but also certain non-local relations with other local structures.

\section{Experiments}
\label{sec:experiments}

\newcommand{\BE}[1]{{\textbf{#1}}}
\begin{table}[t]
	\caption{Comparing the accuracy drop (\%) in 3D shape classification on ModelNet40 between our method and  rotation-variant methods in handling inputs at arbitrary rotations.
		When trained in z and tested in SO3 (i.e., z/SO3), the rotation-variant methods have significant performance drop, compared with z/z (as reference).
		Even we train them in SO3 and test them also in SO3 (i.e., SO3/SO3), their performance still drops considerably. Compared with them, our method has {\em consistent performance\/} when tested on SO3, which validates the rotation invariance of our method.
	}
	\vspace*{-3mm}
	\label{tab:classification_1}
	\centering
	\begin{center}
		\resizebox{1.0\linewidth}{!}{%
			\begin{tabular}{@{\hspace{1mm}}c@{\hspace{1mm}}||@{\hspace{1mm}}C{0.8cm}@{\hspace{1mm}}C{1.2cm}@{\hspace{1mm}}|@{\hspace{1mm}}C{0.9cm}@{\hspace{1mm}}C{1.2cm}@{\hspace{1mm}}|@{\hspace{1mm}}C{1.2cm}@{\hspace{1mm}}}
				\toprule[1pt]
				\multirow{2}*{Method}  & \multirow{2}*{z/SO3} & \multirow{2}*{drop by} & SO3 & \multirow{2}*{drop by} &  z/z \\
				 & && /SO3 && (reference) \\ \hline \hline
				SubVolSup MO~\cite{qi2016volumetric}
				
				& 45.5 & 49.2\% $\hspace{0.3mm}\downarrow$
				& 85.0 & \hspace*{1.4mm}5.0\%  $\hspace{0.3mm}\downarrow$ & 89.5
				\\
				PointNet~\cite{qi2017pointnet}
				
				& 16.4 & 81.6\% $\hspace{0.3mm}\downarrow$
				& 75.5 & 15.4\% $\hspace{0.3mm}\downarrow$ & 89.2
				\\
				PointNet++ (MSG)~\cite{qi2017pointnet++}
				
				& 28.6 & 68.5\% $\hspace{0.3mm}\downarrow$
				& 85.0 & \hspace*{1.4mm}6.3\%  $\hspace{0.3mm}\downarrow$ & 90.7
				\\
				PointCNN~\cite{li2018pointcnn}
				
				& 41.2 & 55.6\% $\hspace{0.3mm}\downarrow$
				& 84.5 & \hspace*{1.4mm}8.6\%  $\hspace{0.3mm}\downarrow$ & 92.5
				\\
				DGCNN~\cite{wang2018dynamic}
				
				& 20.6 & 77.8\% $\hspace{0.3mm}\downarrow$
				& 81.1 & 12.7\% $\hspace{0.3mm}\downarrow$ & 92.9
				\\
				ShellNet~\cite{zhang2019shellnet}
				
				& 19.9 & 78.6\% $\hspace{0.3mm}\downarrow$
				& 87.8 & \hspace*{1.4mm}5.7\%  $\hspace{0.3mm}\downarrow$ & 93.1
				\\
				\hline \hline
				Ours
				
				& \BE{89.4} & \BE{0\%}
				& \BE{89.3} & \hspace*{1.4mm}\BE{0.1\%} $\hspace{0.3mm}\downarrow$ & 89.4
				\\
				\bottomrule[1pt]
		\end{tabular}}
	\end{center}
	\vspace*{-4mm}
\end{table}

\subsection{Implementation Details}
\label{subsec:implementation}

We implemented our network
using TensorFlow~\cite{abadi2016tensorflow} and trained it for 200 epochs in all tasks.
Adam optimizer~\cite{kingma2014adam} was used with a learning rate of 0.001 and a mini-batch size of six.
To avoid over-fitting in the network training, we employed dropout~\cite{hinton2012improving} in our fully-connected layers and also augmented the input point clouds by random scaling and jittering.
Also, we set $N_1$$=$$512$ and $N_2$$=$$128$, and followed~\cite{qi2017pointnet++} to capture multi-scale local regions with different $r$ and $K$ in each layer.
Besides, we empirically set $P_s$$=$$10$ and $K_s$$=$$0.9K$ to balance the computing time and stability in finding the approximate geometric median.
For details on the hyper-parameter settings (\ie, $r$, $K$, and $C$) and time report, please refer to the supplementary material.
Our trained network model and code can be found on the GitHub project page\footnote{\footnotesize{https://github.com/nini-lxz/Rotation-Invariant-Point-Cloud-Analysis}}.

To evaluate the robustness of our network on inputs of arbitrary orientations,
we followed the settings in recent rotation-invariant methods~\cite{zhang2019rotation,chen2019clusternet} to train and test our network in three scenarios:
(i) z/z (as a reference): train and test with rotation augmentation about azimuthal axis,
(ii) z/SO3: train with azimuthal rotations and test with arbitrary rotations, and
(iii) SO3/SO3: train and test both with arbitrary rotations.
Overall, it is expected that an effective rotation-invariant approach should have {\em consistent performance\/} for all scenarios.
In the followings, we evaluate the performance of our method against others, both quantitatively and qualitatively, on three tasks: shape classification (Section~\ref{subsec:classification}), part segmentation (Section~\ref{subsec:segmentation}), and shape retrieval (Section~\ref{subsec:retrieval}).
Then, we show the network component analysis (Section~\ref{subsec:ablation}), various robustness tests (Section~\ref{subsec:noise}), and discuss the limitations (Section~\ref{sec:limit}).

\subsection{Evaluation: 3D Shape Classification}
\label{subsec:classification}
First, we evaluate our method on the 3D shape classification task by comparing it with both rotation-variant and rotation-invariant methods using the standard ModelNet40 dataset~\cite{wu20153d}, which has 12,311 CAD models from the 40 categories.
We adopted the standard split to train our network using 9,843 models and tested it using the remaining 2,468 models.
Note that there is no overlap between the training and test sets.
Each input point cloud has 1024 points.

\para{Comparison with rotation-variant methods.} \
Table~\ref{tab:classification_1} compares the drop in accuracy (\%) for handling inputs at arbitrary rotations.
First, existing rotation-variant methods have significant accuracy drops in z/SO3 as compared with z/z, showing that they are not rotation-invariant.
Second, for the results in SO3/SO3, their performance still drops considerably,
though arbitrary rotations in data augmentation can help improve their performance when testing in SO3.
This means their networks cannot learn in SO3/SO3 as effective as in z/z.
In contrast, the accuracy of our method does not drop for z/SO3, thus demonstrating the rotation invariance of our method.
Also, it outperforms others when tested on SO3, no matter trained on z or on SO3.
Note that, the slight drop in accuracy (\ie, 0.1\%) of our method in SO3/SO3 is caused by the network re-training.
\begin{table}[t]
	\caption{Comparing accuracy drop (\%) in 3D shape classification on ModelNet40 between our method and recent rotation-invariant methods on handling inputs at arbitrary rotations.
		Due to the spherical-related convolutions, SFCNN is not purely rotational invariant, so its performance drops when trained on z and tested on SO3. Methods based on rotation-invariant formulations (RI-ShellConv, ClusterNet, and ours) have a consistent performance, while our method achieves the best performance for testing in SO3.}
	\vspace{-3mm}
	\label{tab:classification_2}
	\centering
	\begin{center}
		\resizebox{1.0\linewidth}{!}{%
			\begin{tabular}{C{2.6cm}||C{1cm}C{1.1cm}|C{1.9cm}}
				\toprule[1pt]
				Method & z/SO3 & drop by &  z/z (reference) \\ \hline \hline
				SFCNN~\cite{rao2019spherical} & 84.8 & 7.2\%$\hspace{1mm}\downarrow$ & 91.4 \\
				RI-ShellConv~\cite{zhang2019rotation} & 86.4 & 0.1\%$\hspace{1mm}\downarrow$ & 86.5 \\
				ClusterNet~\cite{chen2019clusternet} & 87.1 & 0 & 87.1 \% \\ \hline \hline
				Ours & \BE{89.4} & \BE{0\%}  & 89.4  \\
				\bottomrule[1pt]
		\end{tabular}}
	\end{center}
\end{table}

\para{Comparison with rotation-invariant methods.} \
Next, we compare our method with recent rotation-invariant methods.
From the results shown in Table~\ref{tab:classification_2}, we can see that the accuracy of SFCNN drops considerably; since its model relies on spherical-related convolutions and cannot guarantee perfect symmetry in rotations, their results are still sensitive to rotations.
For RI-ShellConv~\cite{zhang2019rotation} and ClusterNet~\cite{chen2019clusternet}, they are formulated with pure rotation invariance, so they have a consistent performance.
Yet, our method still outperforms them, since our rotation-invariant representations encode both local and global information, and our network can effectively learn high-level features more globally with the help of the region relation convolution and global relations.

\begin{table}[t]
	\caption{\new{Comparing 3D shape classification accuracy (\%) on ModelNet40 between our method and recent rotation-equivariant methods. Each method was trained with No Rotations (NR) and tested with NR and Arbitrary Rotations (AR), respectively. Compared with the non-rotation-invariant method PointNet++ (as reference), existing rotation-equivariant methods~\cite{esteves2018learning,zhao2020quaternion,zhang20203d} have significant performance drop when tested on NR, while our method clearly outperforms them and also produces a consistent performance for testing in AR.}}
	\vspace{-3mm}
	\label{tab:classification_3}
	\centering
	\begin{center}
		\resizebox{0.9\linewidth}{!}{%
			\begin{tabular}{C{4.0cm}||C{1cm}C{1cm}}
				\toprule[1pt]
				Method &  NR/NR & NR/AR \\ \hline \hline
				PointNet++~\cite{qi2017pointnet++} (reference) & \BE{89.8} & 21.4 \\ \hline
				Spherical CNN~\cite{esteves2018learning} &- & 43.9 \\
				QE-Capsule Network~\cite{zhao2020quaternion} & 74.4 & 74.1 \\
				REQNN~\cite{zhang20203d} & 83.0 & 83.0 \\
				\hline \hline
				Ours & 89.3 & \BE{89.3} \\
				\bottomrule[1pt]
		\end{tabular}}
	\end{center}
\end{table}

\para{Comparison with rotation-equivariant methods.} \
We also compare our method with recent rotation-equivariant methods~\cite{esteves2018learning,zhao2020quaternion,zhang20203d}.
Specifically, we follow the setting in~\cite{zhao2020quaternion,zhang20203d} to re-train our framework with No Rotation augmentation (denoted as NR), then test it with NR and Arbitrary Rotation (denoted as AR), respectively.
Table~\ref{tab:classification_3} shows the comparison results.
We also provide the results of a non-rotation-invariant network,~\ie, PointNet++, as a reference; see the first row in Table~\ref{tab:classification_3}.
From the table we can see that, these rotation-equivariant methods produce much lower classification accuracies compared with PointNet++ when tested on NR, while our method has only a slight drop.
Further, our method clearly outperforms these rotation-equivariant methods by a significant margin.

\begin{table*}[t]
	\caption{Comparing the object part segmentation performance (per-category mIoU and averaged mIoU) on the ShapeNet dataset for z/SO3 (top) and SO3/SO3 (bottom).}
	\vspace{-3mm}
	\label{tab:seg}
	\centering
	\begin{center}
		\resizebox{1.0\linewidth}{!}{%
			\begin{tabular}{@{\hspace{0.3mm}}c@{\hspace{0.4mm}}|C{0.4cm}@{\hspace{5.0mm}}C{0.6cm}@{\hspace{3.0mm}}C{0.6cm}@{\hspace{3.0mm}}C{0.6cm}@{\hspace{3.0mm}}C{0.7cm}@{\hspace{2.0mm}}C{0.8cm}@{\hspace{2.0mm}}C{0.8cm}@{\hspace{2.2mm}}C{0.7cm}@{\hspace{2.2mm}}C{0.7cm}@{\hspace{1.5mm}}C{0.9cm}@{\hspace{1.5mm}}C{0.8cm}@{\hspace{2.2mm}}C{0.6cm}@{\hspace{2.2mm}}C{0.8cm}@{\hspace{1.5mm}}C{0.9cm}@{\hspace{1.8mm}}C{0.7cm}@{\hspace{2.2mm}}C{0.7cm}@{\hspace{1.5mm}}|@{\hspace{0.6mm}}c@{\hspace{0.3mm}}}
				\toprule[1pt]
				Method (z/SO3) & aero & bag & cap & car & chair & earph. & guitar & knife & lamp & laptop & motor & mug & pistol & rocket & skate & table & avg. mIoU \\ \hline \hline
				PointNet~\cite{qi2017pointnet} & 40.4 & 48.1 &  46.3 &  24.5 &  45.1 &  39.4 &  29.2 &  42.6 &  52.7 &  36.7 &  21.2 &  55.0 &  29.7 &  26.6 &  32.1 &  35.8 &  37.8
				\\
				PointNet++ (MSG)~\cite{qi2017pointnet++}& 51.3 & 66.0 &  50.8 &  25.2 &  66.7 &  27.7 &  29.7 &  65.6 &  59.7 &  70.1 &  17.2 &  67.3 &  49.9 &  23.4 &  43.8 &  57.6 &  48.3\\
				PointCNN~\cite{li2018pointcnn} & 21.8 & 52.0 &  52.1 &  23.6 &  29.4 &  18.2 &  40.7 &  36.9 &  51.1 &  33.1 &  18.9 &  48.0 &  23.0 &  27.7 &  38.6 &  39.9 &  34.7\\
				DGCNN~\cite{wang2018dynamic} & 37.0 & 50.2 &  38.5 &  24.1 &  43.9 &  32.3 &  23.7 &  48.6 &  54.8 &  28.7 &  17.8 &  74.4 &  25.2 &  24.1 &  43.1 &  32.3 &  37.4\\
				ShellNet~\cite{zhang2019shellnet} & 55.8 & 59.4 &  49.6 &  26.5 &  40.3 &  51.2 &  53.8 &  52.8 &  59.2 &  41.8 &  28.9 &  71.4 &  37.9 &  49.1 &  40.9 &  37.3 &  47.2\\ \hline
				PRIN~\cite{you2018pointwise} & 62.6 & 61.2 & 71.1 & 57.0 & 76.7 & 61.8 & 72.3 & 74.7 & 64.7 & 70.6 & 40.1 & 77.0 & 65.0 & 47.8 & 62.3 & 69.4 & 64.6 \\
				RI-ShellConv~\cite{zhang2019rotation} & 80.6 & 80.0 &  70.8 &  68.8 &  86.8 &  70.3 &  87.3 &  \BE{84.7} &  77.8 &  80.6 &  57.4 &  91.2 &  71.5 &  52.3 &  66.5 &  78.4 &  75.3\\ \hline \hline
				Ours & \BE{81.4} & \BE{82.3} &  \BE{86.3} &  \BE{75.3} &  \BE{88.5} &  \BE{72.8} &  \BE{90.3} &  82.1 &  \BE{81.3} &  \BE{81.9} &  \BE{67.5} &  \BE{92.6} &  \BE{75.5} & \BE{54.8}  & \BE{75.1} &  \BE{78.9} &  \BE{79.2} \\
				\bottomrule[1pt]
		\end{tabular}}
	\end{center}
	\vspace{-3mm}
	\begin{center}
		\resizebox{1.0\linewidth}{!}{%
			\begin{tabular}{@{\hspace{0.3mm}}c@{\hspace{0.4mm}}|C{0.4cm}@{\hspace{5.0mm}}C{0.6cm}@{\hspace{3.0mm}}C{0.6cm}@{\hspace{3.0mm}}C{0.6cm}@{\hspace{3.0mm}}C{0.7cm}@{\hspace{2.0mm}}C{0.8cm}@{\hspace{2.0mm}}C{0.8cm}@{\hspace{2.2mm}}C{0.7cm}@{\hspace{2.2mm}}C{0.7cm}@{\hspace{1.5mm}}C{0.9cm}@{\hspace{1.5mm}}C{0.8cm}@{\hspace{2.2mm}}C{0.6cm}@{\hspace{2.2mm}}C{0.8cm}@{\hspace{1.5mm}}C{0.9cm}@{\hspace{1.8mm}}C{0.7cm}@{\hspace{2.2mm}}C{0.7cm}@{\hspace{1.5mm}}|@{\hspace{0.6mm}}c@{\hspace{0.3mm}}}
				\toprule[1pt]
				Method (SO3/SO3) & aero & bag & cap & car & chair & earph. & guitar & knife & lamp & laptop & motor & mug & pistol & rocket & skate & table & avg. mIoU \\ \hline \hline
				PointNet~\cite{qi2017pointnet} & \BE{81.6} & 68.7 &  74.0 &  70.3 &  87.6 &  68.5 &  88.9 &  80.0 &  74.9 &  83.6 &  56.5 &  77.6 &  75.2 &  53.9 &  69.4 &  79.9 &  74.4
				\\
				PointNet++ (MSG)~\cite{qi2017pointnet++}& 79.5 & 71.6 &  \BE{87.7} &  70.7 &  \BE{88.8} &  64.9 &  88.8 &  78.1 &  79.2 &  94.9 &  54.3 &  92.0 &  76.4 &  50.3 &  68.4 &  81.0 &  76.7\\
				PointCNN~\cite{li2018pointcnn} & 78.0 & 80.1 &  78.2 &  68.2 &  81.2 &  70.2 &  82.0 &  70.6 &  68.9 &  80.8 &  48.6 &  77.3 &  63.2 &  50.6 &  63.2 &  \BE{82.0} &  71.4\\
				DGCNN~\cite{wang2018dynamic} & 77.7 & 71.8 &  77.7 &  55.2 &  87.3 &  68.7 &  88.7 &  \BE{85.5} &  \BE{81.8} &  81.3 &  36.2 &  86.0 &  \BE{77.3} &  51.6 &  65.3 &  80.2 &  73.3\\
				ShellNet~\cite{zhang2019shellnet} & 79.0 & 79.6 &  80.2 &  64.1 &  87.4 &  71.3 &  88.8 &  81.9 &  79.1 &  \BE{95.1} &  57.2 &  91.2 &  69.8 &  \BE{55.8} &  73.0 &  79.3 &  77.1\\ \hline
				PRIN~\cite{you2018pointwise} & 67.4 & 61.5 & 69.7 & 59.5 & 77.7 & 65.8 & 75.7 &77.2 & 65.9 & 82.0 & 44.2 & 79.8 & 63.6 & 53.0 & 67.5 & 70.7 & 67.6 \\
				RI-ShellConv~\cite{zhang2019rotation} & 80.6 & 80.2 &  70.7 &  68.8 &  86.8 &  70.4 &  87.2 &  84.3 &  78.0 &  80.1 &  57.3 &  91.2 &  71.3 &  52.1 &  66.6 &  78.5 &  75.3\\ \hline \hline
				Ours &  81.4 & \BE{84.5} &  85.1 &  \BE{75.0} &  88.2 &  \BE{72.4} &  \BE{90.7} &  84.4 &  80.3 &  84.0 &  \BE{68.8} &  \BE{92.6} &  76.1 &  52.1 &  \BE{74.1} &  80.0 &  \BE{79.4} \\
				\bottomrule[1pt]
		\end{tabular}}
	\end{center}
\end{table*}

\subsection{Evaluation: 3D Object Part Segmentation}
\label{subsec:segmentation}

Next, we evaluate our method on part segmentation by comparing it with both rotation-variant methods and the recent rotation-invariant methods PRIN~\cite{you2018pointwise} and RI-ShellConv~\cite{zhang2019rotation} on the ShapeNet dataset~\cite{chang2015shapenet}.
This dataset has 16,881 models from 16 categories, and is annotated with 50 parts.
Also, there is no overlap between the training and testing sets.
For both PRIN and RI-ShellConv, we directly used their released code to re-train their networks.
We adopt the per-category averaged intersection over union (mIoU) metric~\cite{qi2017pointnet} in the evaluation.
Here, we do not compare with ClusterNet~\cite{chen2019clusternet}, since it is designed for classification.

\begin{figure}[!t]
	\centering
	\includegraphics[width=0.95\linewidth]{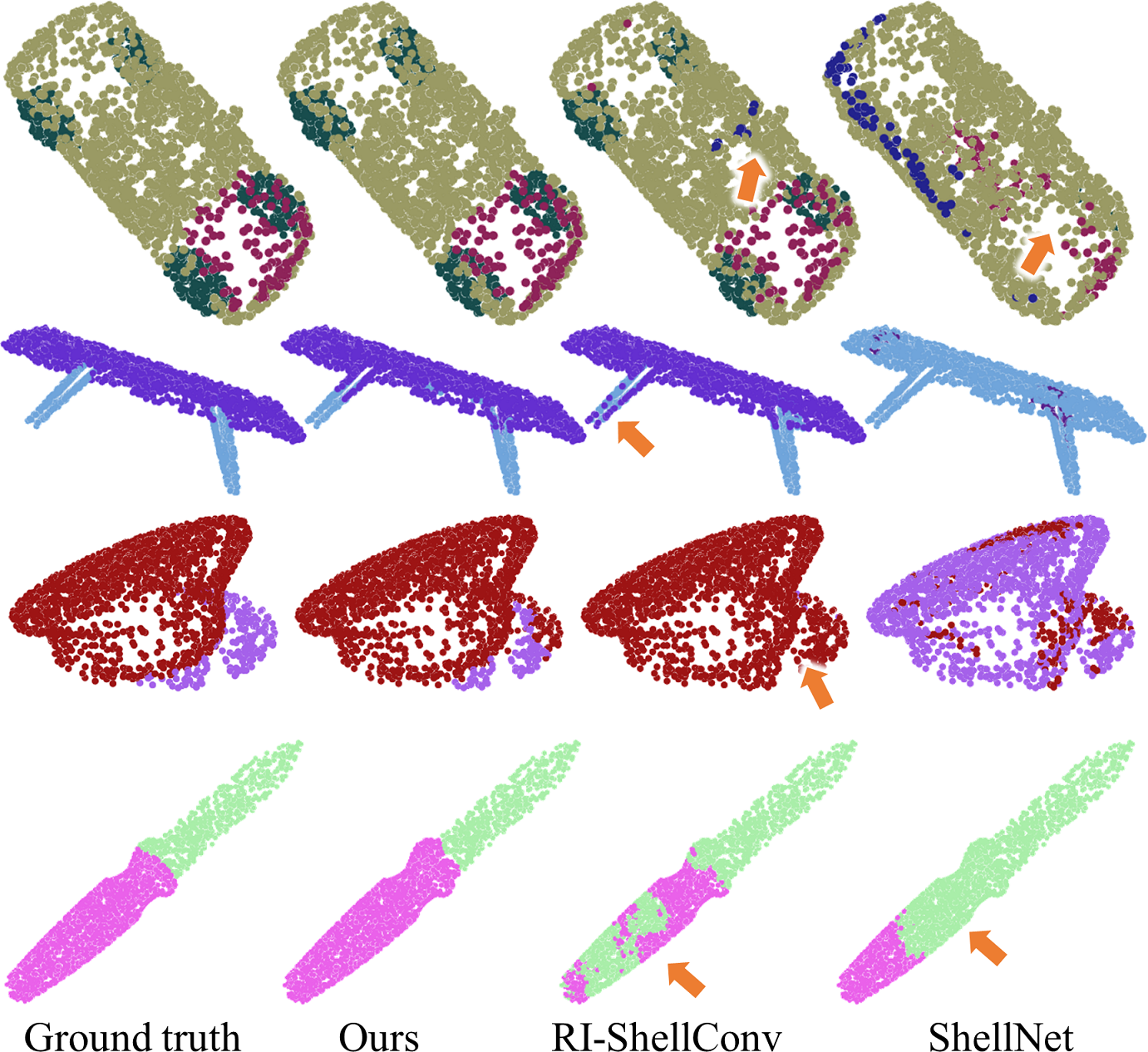}
	\vspace{-1mm}
	\caption{Visual comparison results on object part segmentation in the z/SO3 scenario.}
	\label{fig:seg_vis}
	\vspace{-3mm}
\end{figure}

Table~\ref{tab:seg} shows the per-category mIoU and averaged mIoU (across all 16 categories) produced by different methods in scenarios z/SO3 and SO3/SO3.
Comparing the results shown in the top and bottom tables, we can see that the rotation-variant methods yield very different segmentation results for z/SO3 and SO3/SO3, while both RI-ShellConv and our method achieve more consistent performance when tested on inputs at arbitrary rotations.
Also, our method outperforms RI-ShellConv and others with the highest averaged mIoU; see the right-most columns in the two tables.
Again, due to the network re-training, although both RI-ShellConv and our method are rotation invariant, there are slight difference in the results for z/SO3 and SO3/SO3.
Further, we show some typical visual comparison results for the z/SO3 scenario in Figure~\ref{fig:seg_vis}, where the segmentation results produced by our method are the closest to the ground truths, compared with others.
Please see supplementary material for more visual comparisons.

\begin{table*}[t]
	\caption{Comparing the 3D shape retrieval performance of our method with the state-of-the-arts on the perturbed dataset in the SHREC'17 contest.
		Overall, a larger evaluation metric value (P@N, R@N, etc.) indicates a better performance.
		Please refer to~\cite{savva2017shrec17} for the detail of each metric, and also the micro and macro average strategies.}
	\vspace{-2.5mm}
	\label{tab:retrieval}
	\centering
	\begin{center}
		\resizebox{0.95\linewidth}{!}{%
			\begin{tabular}{c||ccccc|ccccc||c}
				\toprule[1pt]
				\multirow{2}*{Method} & \multicolumn{5}{c|}{micro} & \multicolumn{5}{c||}{macro} & \multirow{2}*{avg}
				\\ 
				& P@N & R@N & F1@N & mAP & NDCG & P@N & R@N & F1@N & mAP & NDCG & \\ \hline \hline
				Furuya~\cite{furuya2016deep} (\emph{contest winner}) & 0.814 & 0.683 & 0.706 & 0.656 & 0.754 & 0.607 & 0.539 & 0.503 & 0.476 & 0.560 & 0.566
				\\ \hline
				SFCNN~\cite{rao2019spherical} & 0.778 & \BE{0.751} & \BE{0.752} & 0.705 & \BE{0.813} & \BE{0.656} & 0.539 & 0.536 & 0.483 & \BE{0.580} & 0.594\\
				RI-ShellConv~\cite{zhang2019rotation} & 0.641 & 0.698 & 0.639 & 0.606 & 0.721 & 0.325 & 0.608 & 0.368 & 0.445 & 0.531 & 0.526\\ \hline \hline
				Ours & \BE{0.821} & 0.737 & 0.741 & \BE{0.707} & 0.805 & 0.512 & \BE{0.664} & \BE{0.558} & \BE{0.510} & 0.560 & \BE{0.609}\\
				\bottomrule[1pt]
		\end{tabular}}
	\end{center}
\end{table*}
\subsection{Evaluation: 3D Shape Retrieval}
\label{subsec:retrieval}

Besides 3D shape classification and object part segmentation, we further evaluate our method on 3D shape retrieval using the perturbed ShapeNet Core55 dataset~\cite{chang2015shapenet}.
Here, we followed the rules of the SHREC'17 3D shape retrieval contest~\cite{savva2017shrec17}, where each model has been randomly rotated by a uniformly-sampled rotation in SO(3).
For a fair comparison, we trained and tested all methods on the provided training/validation/testing sets, and evaluated their performance with the official evaluation metrics,~\ie, precision (P@N), recall (R@N), F1-score (F1@N), mean average precision (mAP), and normalized discounted cumulative gain (NDCG).
On each shape, 2048 points were sampled as the network input.
To combine the retrieval results of different categories, we followed~\cite{savva2017shrec17} to use the macro and micro average strategies on the above five metrics.
\new{For a better demonstration, we also followed the official SHREC'17 contest to compute the average score between mAPs by using the micro and macro strategies as another metric.}

Table~\ref{tab:retrieval} reports the evaluation results.
Overall, a larger metric value indicates a better retrieval performance.
Compared with the contest winner~\cite{furuya2016deep} and also the recent rotation-invariant methods~\cite{rao2019spherical,zhang2019rotation}, our method achieves comparable performance on most evaluation metrics, and also the best average score (see the right-most column) by a large margin compared with others.


\subsection{Method Component Analysis}
\label{subsec:ablation}

\begin{table*}[t]
	\caption{Ablation study, representation analysis, and network architecture analysis (left to right) using the shape classification accuracy on ModelNet40 in z/SO3.}
	\vspace{-2.5mm}
	\label{tab:ablation}
	\centering
	\begin{center}
		\resizebox{0.99\linewidth}{!}{%
			\begin{tabular}{C{1.2cm}||C{1.1cm}C{1.1cm}C{1.1cm}|C{3.5cm}|C{1.7cm}C{1.7cm}||C{1.7cm}}
				\toprule[1pt]
				\multirow{2}*{Scenario}& \multicolumn{3}{c|}{Ablation study} & Rot.-inv. representation & \multicolumn{2}{c||}{Network architecture} & \multirow{2}*{Full pipeline}\\ \cline{2-7}
				& 	Case \#1 & Case \#2 & Case \#3 & RI-ShellConv~\cite{zhang2019rotation} & PointNet++ & DGCNN & \\ \hline
				z/SO3 & 88.4 & 88.1 & 88.4 & 87.8 & 88.6 & 82.6 & \BE{89.4} \\				
				\bottomrule[1pt]
		\end{tabular}}
	\end{center}
\end{table*}

Next, we present an ablation study, an analysis on our rotation-invariant representation, and a network architecture analysis to evaluate different aspects of our method using the shape classification task on ModelNet40.

\para{Ablation study.} \
First, we evaluate the following three major modules in our method:
\begin{itemize}
%
\item
\emph{Case \#1}.
To evaluate the effectiveness of our designed global rotation-invariant representation $G_i$ (see Eq.~\eqref{equ:center}), we remove it and keep only the local representations $L_{ij}$.
Hence, for the matrix (network inputs) presented in Figure~\ref{fig:matrix}, we only keep the right-half,~\ie, $K \times 7$, to characterize each point $p_i$; and
%
\item
\emph{Case \#2}.
\new{To evaluate the effectiveness of our proposed local rotation-invariant representation $L_{ij}$ (see Eq.~\eqref{equ:neighbor_all}), we replace it with the representation in RI-ShellConv~\cite{zhang2019rotation}, while keeping our global rotation-invariant representation; and}
\item
\emph{Case \#3}.
To evaluate the effectiveness of our proposed region relation convolution, we remove the upper branch in region relation convolution and directly take $\mathcal{F}$ as the output (see Figure~\ref{fig:conv}).
\end{itemize}

\vspace{-1mm}
The leftmost portion of Table~\ref{tab:ablation} shows the results of the three cases.
Since all the three cases are rotation invariant, their classification accuracies are consistent for z/z, z/SO3, and SO3/SO3, so we report only the accuracies for z/SO3 in the table.
By comparing the result with our full pipeline (rightmost in Table~\ref{tab:ablation}), we can see that each module (case) contributes to achieve a better classification performance.

\vspace{1mm}
\para{Rotation-invariant representation analysis.} \
To verify the effectiveness of our rotation-invariant representation (both $G_i$ and $L_{ij}$, as depicted in Figure~\ref{fig:matrix}), we replace it entirely with the state-of-the-art rotation-invariant representation proposed in~\cite{zhang2019rotation}.
The resulting classification accuracy is shown in the middle portion of Table~\ref{tab:ablation}.
Comparing with our full-pipeline result (rightmost in Table~\ref{tab:ablation}), we can see that our network achieves a better performance with our rotation-invariant representation (89.4\%) than with the representation in~\cite{zhang2019rotation} (87.8\%).
However, such performance (87.8\%) is still higher than the performance (86.4\%) of~\cite{zhang2019rotation} (see Table~\ref{tab:classification_2}).
The difference reveals that while both cases use the same rotation-invariant representation from~\cite{zhang2019rotation}, our network with the region relation convolution and global relation information can achieve a better performance.

\vspace{1mm}
\para{Network architecture analysis.} \
To verify the effectiveness of our network (Figure~\ref{fig:network}), we replace it with PointNet++~\cite{qi2017pointnet++} and DGCNN~\cite{wang2018dynamic}, while keeping our rotation-invariant representation as the network input.
The ``network architecture'' column in Table~\ref{tab:ablation} shows the results.
Apparently, our network (full pipeline) achieves a higher performance.
Also, we explore the performance of our network with different number of layers; see the supplementary material.


\subsection{Robustness Test}
\label{subsec:noise}

\begin{figure}[t]
	\centering
	\includegraphics[width=0.8\linewidth]{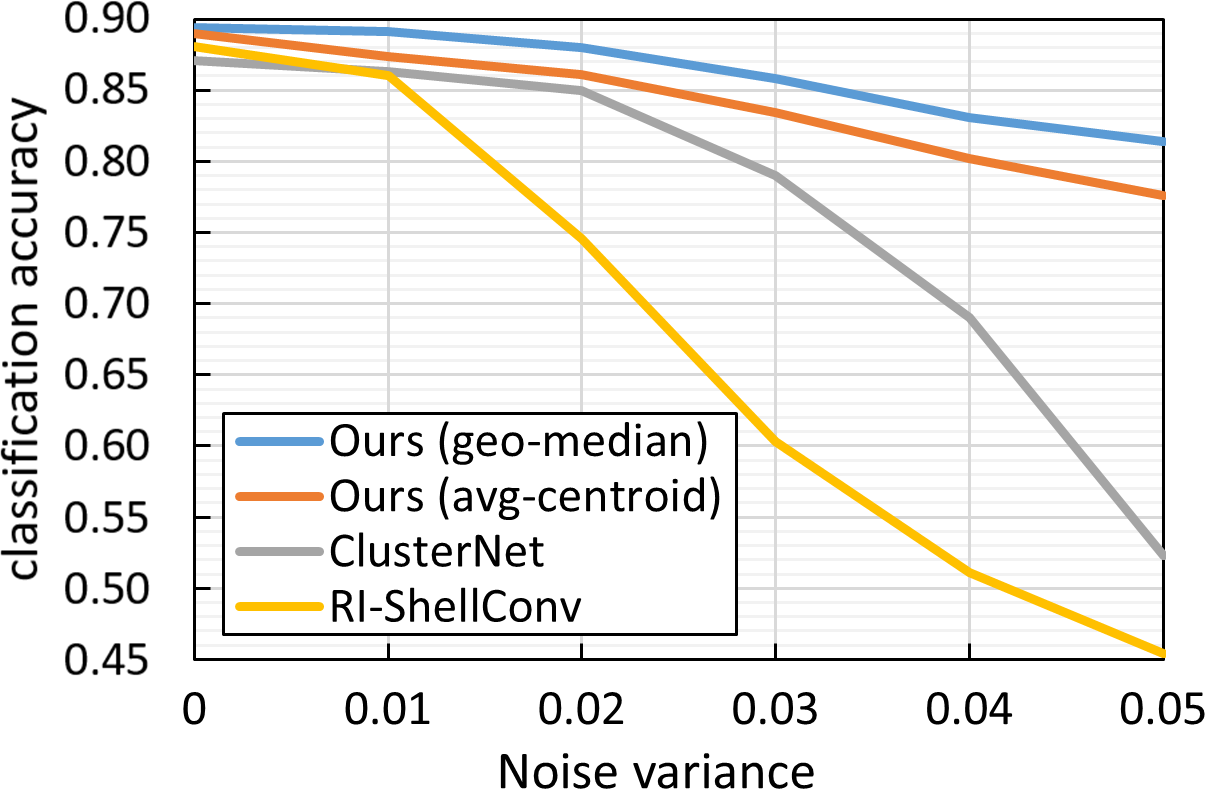}
	\caption{Comparing the shape classification accuracy of various methods on ModelNet40, for inputs corrupted by Gaussian noise of increasing noise level (variance).}
	\label{fig:noise}
	\vspace{-3mm}
\end{figure}

\para{Robustness on noise.} \
Noise is common in the acquisition of 3D point clouds.
This motivates us to introduce the geometric median (which is less sensitive to noise) for formulating our rotation-invariant representations.
To study our method's robustness to noise, we test its shape classification performance on ModelNet40 using inputs that are corrupted by Gaussian noise of increasing level (variance).
In this test, we consider four cases:
(i) our method with geometric median;
(ii) our method with arithmetic mean;
(iii) ClusterNet~\cite{chen2019clusternet}; and
(iv) RI-ShellConv~\cite{zhang2019rotation}.

Figure~\ref{fig:noise} plots the shape classification accuracy for the four cases using input shapes of increasing amount of noise.
For ClusterNet, their authors kindly help us generate the results.
For RI-ShellConv, we use their released code to produce the results.
From the results, we can see that our method with geometric median consistently achieves better performance than with the arithmetic mean, and existing rotation-invariant methods are more sensitive to noise.

\begin{figure}[t]
	\centering
	\includegraphics[width=1.0\linewidth]{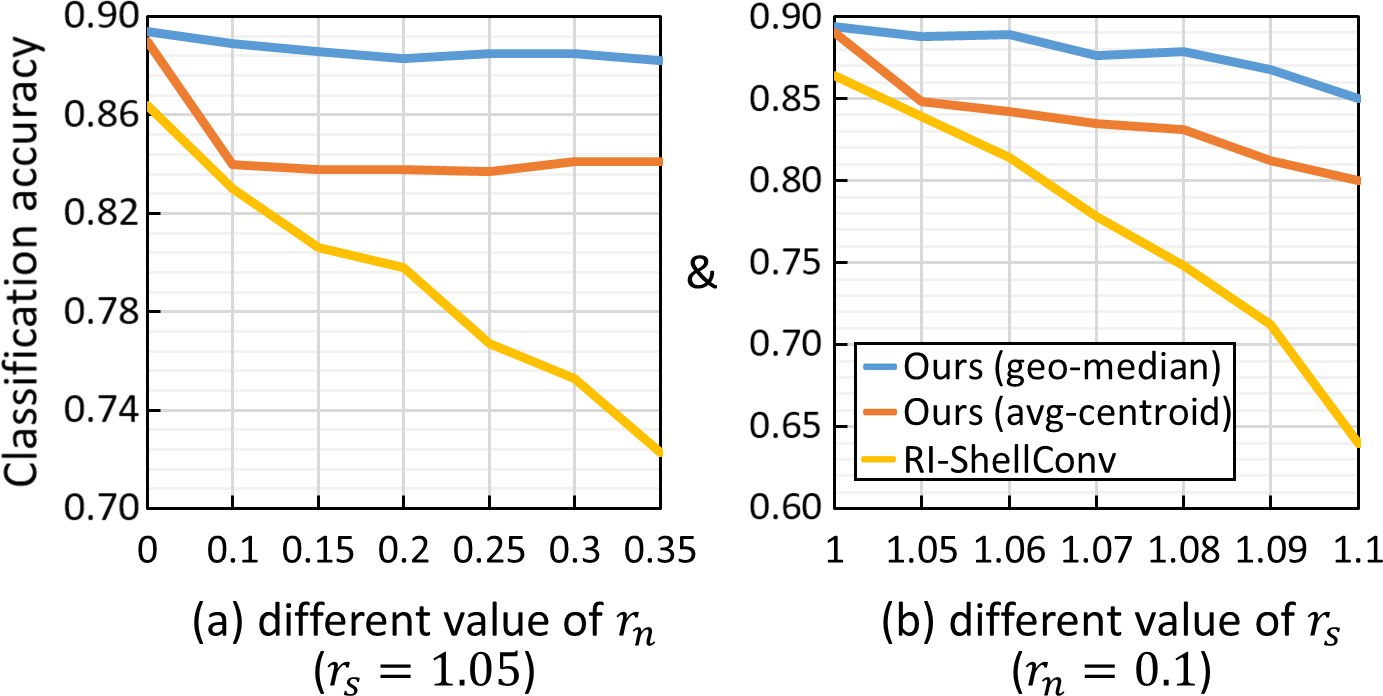}
	\caption{Comparing the shape classification accuracy of various cases on ModelNet40, for inputs contaminated with outliers of increasing $r_n$ (a) and $r_s$ (b); see Section~\ref{subsec:noise} for the detailed explanations of $r_n$ and $r_s$.}
	\label{fig:outlier}
	\vspace{-3mm}
\end{figure}

\para{Robustness on outliers.} \
Next, we study the robustness of our method on handling inputs contaminated with outliers.
Similar to the noise robustness test, we also consider cases of (i) our method with geometric median, (ii) our method with arithmetic mean, and (iii) RI-ShellConv~\cite{zhang2019rotation}.
Note that ClusterNet~\cite{chen2019clusternet} is not included in this experiment, since there is no publicly available code and we did not receive feedback from the authors.
In this experiment, we test the shape classification performance of the three cases on ModelNet40 using inputs with extra outliers.
For each test shape of $N$ points ($N$$=$$1024$ on ModelNet40), we add $N$$\times$$r_n$ outliers, where $r_n$ is a ratio that controls the number of outliers.
To generate outliers, we randomly added points within a sphere of radius $r_s$ co-centered with the given shape (which has been normalized to fit a unit sphere), while avoiding the outlier points to be too close to any existing data point; the threshold distance here is empirically set to 5\% of the input shape size. Intuitively, the larger the values of $r_n$ and $r_s$ are, the more severe the outliers will be.

Figures~\ref{fig:outlier} (a)\&(b) plot the shape classification accuracy for inputs of increasing $r_n$ and $r_s$, respectively.
For the case of Figure~\ref{fig:outlier} (a), when $r_n=0$, the test inputs are clean without outliers. For the the case of Figure~\ref{fig:outlier} (b), we set $r_n=0$ when $r_s=1$ and set $r_n=0.1$ when $r_s>1$, so that we can show the accuracy for an outlier-free condition as a reference in the plot. For both cases, we directly employ the network model (which is previously-trained on ModelNet40) to test on inputs contaminated with extra outlier points, without re-training the network models. Clearly, when compared with other methods, our method with geometric median (blue plot) consistently achieves better and also more stable classification performance.

\para{Robustness on varying input density.} \
Third, we test the robustness of our method on handling inputs with varying density.
Specifically, for shapes in both the training and test sets of ModelNet40, we first scale them, such that all have the same surface area, and sample different number of points on each shape to obtain point sets of varying densities. In detail, we prepared three sets of point sets of increasing densities,~\ie, with 512, 1024, and 2048 points, respectively. Then, we compared our method against the recent rotation-invariant method RI-ShellConv~\cite{zhang2019rotation} on the shape classification task. For each method, we re-trained the network using each point set in the training set, and then tested the trained network models on the point sets of the associated test set to obtain the classification accuracies under the z/SO3 scenario. Results show that, when using training shapes with 512, 1024, and 2048 points (increasing density), our method achieves accuracies of 87.8\%, 89.4\%, and 89.3\%, respectively, whereas RI-ShellConv achieves accuracies of 85.1\%, 86.4\%, and 86.7\%, respectively. Hence, our method consistently has higher accuracy than RI-ShellConv for input point sets of different densities.

\begin{figure}[t]
	\centering
	\includegraphics[width=1.0\linewidth]{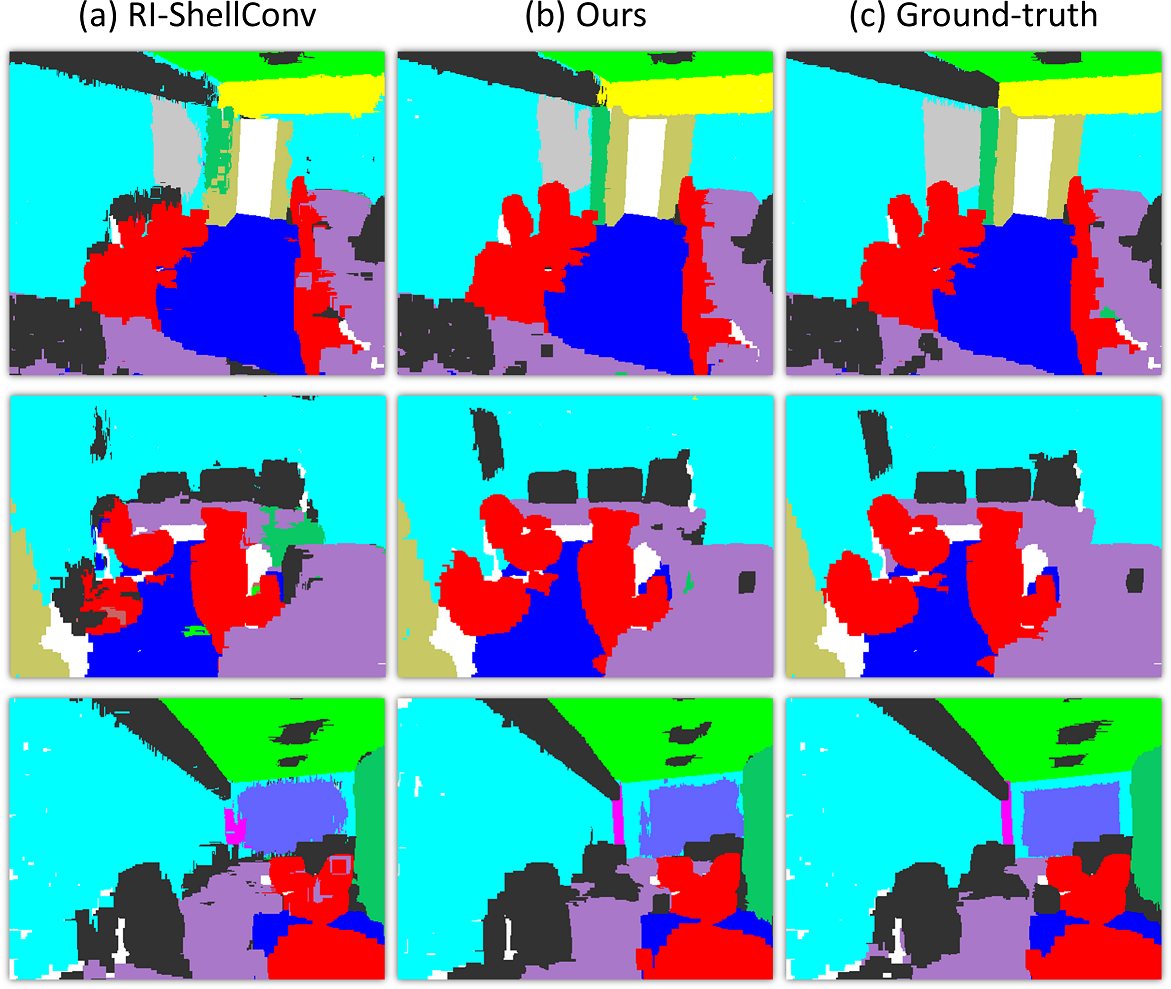}
	\caption{Comparing the semantic scene segmentation results by using our method (b) and RI-ShellConv (a) against the ground-truth (c) on Area 6 of S3DIS dataset.}
	\label{fig:sem}
	\vspace{-3mm}
\end{figure}

\para{Robustness on semantic scene segmentation.} \
Further, we tested the robustness of our method on semantic scene segmentation.
In detail, we conducted the experiment on the commonly-used Stanford 3D semantic parsing (S3DIS) dataset~\cite{armeni20163d}, which contains 3D scans in 6 Areas with 271 rooms altogether. We followed the common practice in many works, including PointNet, to train both our network and RI-ShellConv on Areas 1-5 without rotation augmentation and then test the segmentation performance on Area 6 with arbitrary rotation augmentation. The per-point overall accuracy achieved by our method is 89.0\%, compared with 80.7\% of RI-ShellConv. 
Figure~\ref{fig:sem} further shows some qualitative comparison cases. Our segmentation results (b) are closer to the ground truths (c).

\para{Robustness on real-scanned incomplete inputs.} \
Lastly, we tested the robustness of our method on handling real-scanned incomplete inputs.
Here, we use ScanObjectNN~\cite{uy2019revisiting}, a recent real-scanned dataset that contains 2902 incomplete point cloud objects in 15 categories. Note that the point sets in ScanObjectNN may contain noise, object partiality, non-uniform point distribution, and even deformation variants. We followed the official train/test split to re-train both our network and RI-ShellConv on the training set of ScanObjectNN and then test the classification performance on the testing set of ScanObjectNN under the z/SO3 scenario. Results show that the classification accuracies on the testing set are 79.8\% (Ours) vs. 78.0\% (RI-ShellConv), demonstrating that our method can also achieve better performance on the ScanObjectNN dataset.

\subsection{Limitations}
\label{sec:limit}
Although our method achieves promising performance against the state-of-the-arts, there are still some limitations.
First, compared with absolute 3D point coordinates, though we have tried to avoid information loss when designing our rotation-invariant representations (\ie, Eqs.~\eqref{equ:center} \& \eqref{equ:neighbor_all}), a small amount of information loss is still inevitable, since we only rely on $L_2$ distances and relative angles, thus resulting in a slightly lower performance in the z/z scenario compared with the existing non-rotation-invariant methods; see Table~\ref{tab:classification_1}.
Second, despite the effectiveness of our method (see Figure~\ref{fig:noise}) on handling noisy inputs as compared with others, the performance still drops progressively when the noise is too large.
So, a network dedicated for noise robustness may be required, since noise is common in real-scanned point clouds.
Lastly, there are different invariance scenarios,~\eg, global rotation of a whole model, local rotation (or deformation) of a region, and permutation of points; our current approach handles mainly the global rotation.

\section{Conclusion}
\label{sec:conclusion}
We presented a rotation-invariant framework for deep 3D point cloud analysis.
Given an input cloud at arbitrary orientation, our framework produces consistent, and also the best performance, on multiple point cloud analysis tasks, including shape classification, part segmentation, and shape retrieval, compared with the state-of-the-arts.
To achieve this, we introduce a novel low-level purely rotation-invariant representation as the network inputs, which encodes both local and global information, as well as being robust to noise and outliers.
Further, we formulate the region relation convolution to enrich the network features with more global information.
The extensive experimental results confirm the rotation invariance of our method, and also its superiority over the state-of-the-arts.

In the future, we plan to explore the possibility of designing a noise-resistant network,
since LiDAR-scanned real inputs are often contaminated with a large amount of noise, particularly for outdoor situations.
Besides, we plan also to extend our rotation-invariant framework for the problems of point cloud registration and partial shape matching, in which, we aim to consider the rotation invariance of local regions instead of the whole input.
Further, we plan to extend our framework to handle large-scale 3D point clouds to help detect or locate object proposals with rotation invariance.
Another possible direction is to explore domain adaptation techniques to adapt our current approach to handle real-scanned objects, such as the objects in the Redwood dataset~\cite{Choi2016}.

\section*{Acknowledgments}
We thank reviewers for their valuable comments.
The work was supported by the Hong Kong Centre for Logistics Robotics, Hong Kong Research Grants Council with Project No. CUHK 14206320 \& 14201620, and the National Natural Science Foundation of China (Project No. 62006219).
This work was also supported in part by the Israel Science Foundation (grant no 2492/20).

\bibliographystyle{IEEEtran}
\bibliography{egbib}
%
%
\begin{IEEEbiography}[{\includegraphics[width=1in,height=1.25in,clip,keepaspectratio]{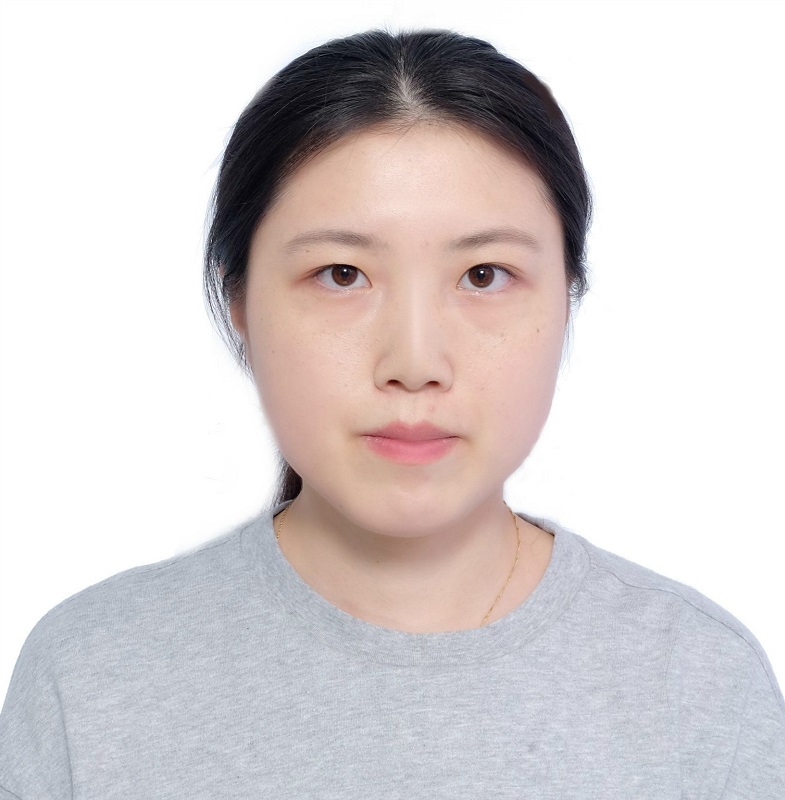}}]{Xianzhi Li}
	is currently a post-doctoral fellow in the Chinese University of Hong Kong. She received her Ph.D. degree in the Department of Computer Science and Engineering from the Chinese University of Hong Kong. She serves as the reviewer of several conferences and journals, including TVCG, CVPR, ICCV, etc. Her research interests focus on 3D vision, computer graphics, and deep learning.
\end{IEEEbiography}
\vspace{-10mm}
\begin{IEEEbiography}[{\includegraphics[width=1in,height=1.25in,clip,keepaspectratio]{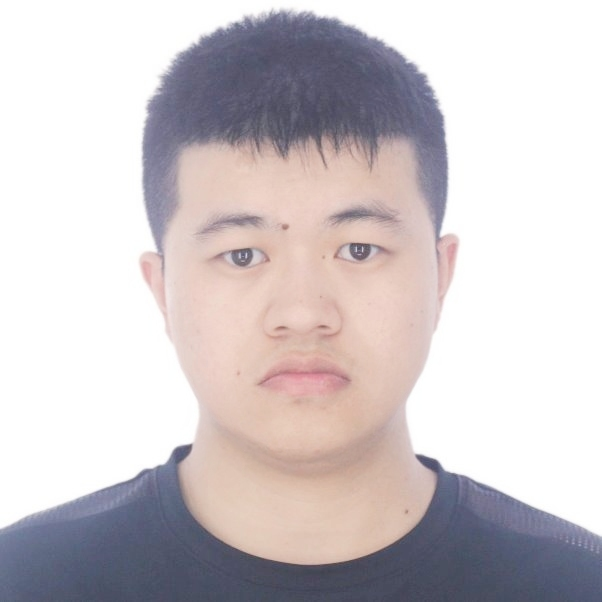}}]{Ruihui Li} received his Ph.D. degree in the Department of Computer Science and Engineering from the Chinese University of Hong Kong in 2021. He serves as the reviewer of several conferences and journals, including TPAMI, CVPR, ICCV, etc. His research interests include 3D vision, geometric processing and modeling, computer graphics, and deep learning.
\end{IEEEbiography}
\vspace{-10mm}
\begin{IEEEbiography}[{\includegraphics[width=1in,height=1.25in,clip,keepaspectratio]{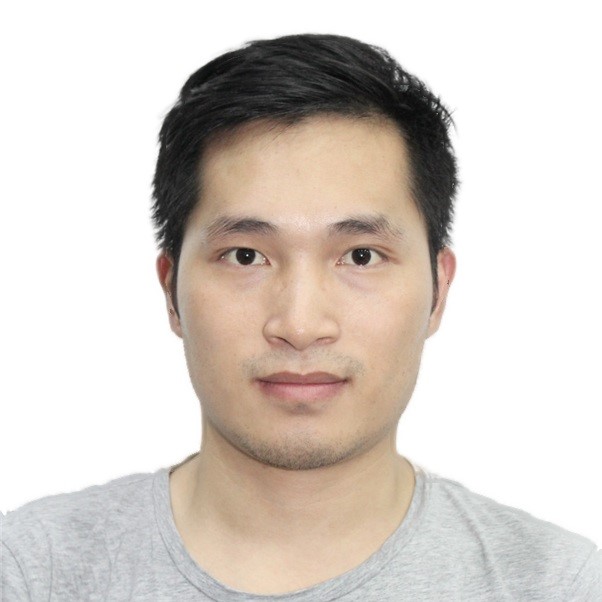}}]{Guangyong Chen}
	is currently an associate researcher in the Shenzhen Insititutes of Advanced Technology, Chinese Academic of Science. He achieved his B.Sc. from Nanjing University in 2012 and then his PhD from the Chinese University of Hong Kong in 2016. Guangyong Chen serves as the reviewer and PC member of several conferences and journals, including TVCG, TOG, ICML, TNNLS, TKDD, CVPR, ICCV, ECCV, etc. His recent research interests include the fundamental of machine learning with its applications in computer vision, computer graphics, etc.
\end{IEEEbiography}
\vspace{-10mm}
\begin{IEEEbiography}[{\includegraphics[width=1.0in,height=1.25in,clip,keepaspectratio]{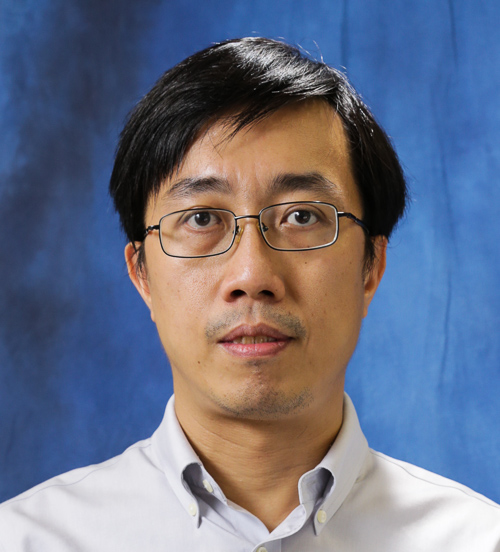}}]{Chi-Wing Fu} is currently an associate professor in the Chinese University of Hong Kong.  He served as the co-chair of SIGGRAPH ASIA 2016's Technical Brief and Poster program, associate editor of IEEE Computer Graphics \& Applications and Computer Graphics Forum, panel member in SIGGRAPH 2019 Doctoral Consortium, and program committee members in various research conferences, including SIGGRAPH Asia Technical Brief, SIGGRAPH Asia Emerging tech., IEEE visualization, CVPR, IEEE VR, VRST, Pacific Graphics, GMP, etc.  His recent research interests include computation fabrication, point cloud processing, 3D computer vision, user interaction, and data visualization.
\end{IEEEbiography}
\vspace{-10mm}
\begin{IEEEbiography}[{\includegraphics[width=1.0in,height=1.25in,clip,keepaspectratio]{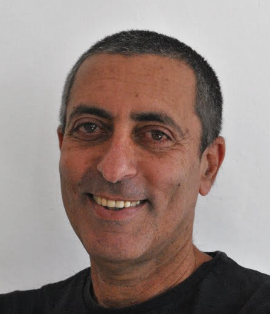}}]{Daniel-Cohen Or} is a professor in the School of Computer Science, Tel Aviv University. He received his B.Sc. cum laude in both mathematics and computer science (1985), and M.Sc. cum laude in computer science (1986) from Ben-Gurion University, and Ph.D. from the Department of Computer Science (1991) at State University of New York at Stony Brook. He was sitting on the editorial board of a number of international journals, and a member of many the program committees of several international conferences. He was the recipient of the Eurographics Outstanding Technical Contributions Award in 2005, ACM SIGGRAPH Computer Graphics Achievement Award in 2018.
	In 2013 he received The People’s Republic of China Friendship Award. In 2015 he has been named a Thomson Reuters Highly Cited Researcher. In 2019 he won The Kadar Family Award for Outstanding Research. In 2020 he received The Eurographics Distinguished Career Award.
	His research interests are in computer graphics, in particular, synthesis, processing and modeling techniques.  His main interest right now is in few areas: image synthesis, motion and transformations, shapes, surfaces, analysis and reconstruction and information visualization.
\end{IEEEbiography}
\vspace{-10mm}
\begin{IEEEbiography}[{\includegraphics[width=1in,height=1.25in,clip,keepaspectratio]{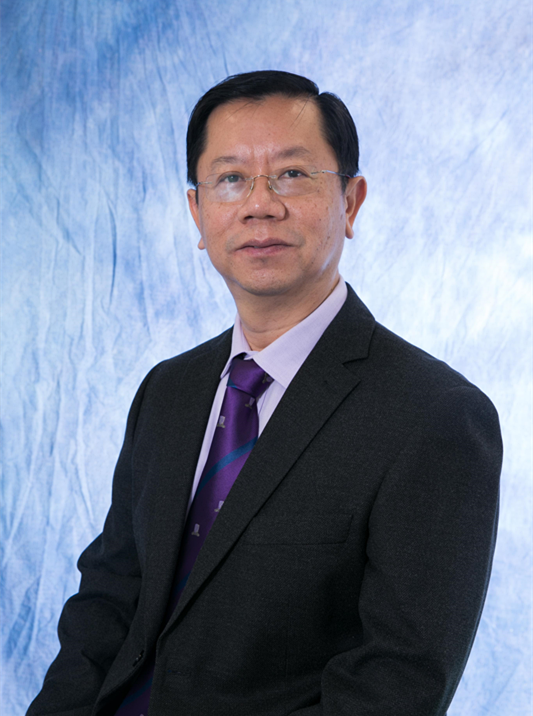}}]{Pheng-Ann Heng} received his B.Sc. from the National University of Singapore in 1985. He received his MSc (Comp. Science), M. Art (Applied Math) and Ph. D (Comp. Science) all from the Indiana University of USA in 1987, 1988, 1992 respectively. He is a professor at the Department of Computer Science and Engineering at The Chinese University of Hong Kong (CUHK). He has served as the Director of Virtual Reality, Visualization and Imaging Research Center at CUHK since 1999 and as the Director of Center for Human-Computer Interaction at Shenzhen Institute of Advanced Integration Technology, Chinese Academy of Science/CUHK since 2006. He has been appointed as a visiting professor at the Institute of Computing Technology, Chinese Academy of Sciences as well as a Cheung Kong Scholar Chair Professor by Ministry of Education and University of Electronic Science and Technology of China in 2007. His research interests include AI and VR for medical applications, surgical simulation, visualization, graphics and human-computer interaction.
\end{IEEEbiography}







\end{document}